\documentclass[conference]{IEEEtran}
\IEEEoverridecommandlockouts
\usepackage{cite}
\usepackage{amsmath,amssymb,amsfonts}
\usepackage{algorithm}
\usepackage{algorithmic}
\usepackage{graphicx}
\usepackage[font=small]{caption}
\usepackage[font=footnotesize]{subcaption}
\usepackage{textcomp}
\usepackage{hyperref}
\usepackage{xcolor}
\newcommand{\rev}[1]{\textcolor{black}{#1}}
\hypersetup{
    colorlinks=true,
    linkcolor=blue,
    filecolor=magenta,      
    urlcolor=cyan,
    pdftitle={Overleaf Example},
    pdfpagemode=FullScreen,
    }
\def\BibTeX{{\rm B\kern-.05em{\sc i\kern-.025em b}\kern-.08em
    T\kern-.1667em\lower.7ex\hbox{E}\kern-.125emX}}
\begin{document}

\title{A Modular Dual-Camera Pipeline for Micro-Inspection Using Aerial Robots\\
{\footnotesize}
\thanks{This work was partly supported by the Netherlands Organization for Scientific Research (NWO) via SIA RAAK-Public project (Van bestrijden naar beheersen van de EPR, No.10.015) and SIA RAAK-MKB project (Smart Greenhouses, No.17.014).}
}

\author{%
\IEEEauthorblockN{Seyed Hojat Mirtajadini$^{1}$, Nils Rublein$^{2}$, Rahul Moongayil Ramakrishnan$^{3}$,\\
Gerjen ter Maat$^{4}$, Mohammad Aldibaja$^{5}$, Abeje Yenehun Mersha$^{6}$}
\IEEEauthorblockA{$^{1,2,3,4,6}$\textit{Smart Mechatronics and Robotics (SMART) Research Group},\\
\textit{Saxion University of Applied Sciences}, Enschede, The Netherlands \\
$^{5}$\textit{Faculty of Science and Engineering},
\textit{University of Groningen}, The Netherlands \\
$^{1}$s.h.mirtajadinigoki@saxion.nl,
$^{2}$n.rublein@saxion.nl,
$^{3}$r.moongayilramakrishnan@saxion.nl,\\
$^{4}$g.j.termaat@saxion.nl,
$^{5}$m.aldibaja@rug.nl,
$^{6}$a.y.mersha@saxion.nl}
}

\maketitle

\begin{abstract}
Most existing drone-based inspection systems require the drone to fly dangerously close to the target or follow complex flight paths to capture small details. 
In addition, drone flight is affected by disturbances and localization inaccuracies, which can cause the drone to lose sight of its supposed target when it has a narrow view. 
Furthermore, trajectory planning often requires prior information about the target's geometry, position, and orientation, which is not always available for non-structural targets such as trees, vehicles, or people.
To address these challenges, this paper presents \texttt{aerial\_micro\_inspection}, a generic pipeline for aerial micro-inspection across different use cases. 
The pipeline assumes a PX4-powered drone equipped with two cameras: (i) a zoomed, gimbal-mounted inspection camera that captures fine details without requiring the drone to fly very close to the target, and (ii) a wide-field-of-view stereo navigation camera that acquires the target surface on site, estimates its range, and partitions it into smaller inspection regions. In addition, a vision-based feedback loop compensates for drone motion while the inspection camera visits small partitions of a larger surface.
We evaluate the pipeline in simulation and real-world experiments, mainly in two use-case scenarios: tree inspection for detecting oak processionary caterpillars and their eggs, and greenhouse inspection of sticky traps for detecting whiteflies. 
The results show improved coverage robustness under drone disturbances in simulation, as well as effective detection of caterpillars and eggs and high-detail imaging of insects in real-world experiments. 
The pipeline is open-source, developed in ROS~2, and can be adapted to new applications by replacing the surface-segmentation and micro-target detection checkpoints.
The code is available at:
\url{https://github.com/SaxionMechatronics/aerial_micro_inspection}
\end{abstract}

\begin{IEEEkeywords}
aerial robot, visual inspection, gimbal control, \rev{visual servoing}, calibration
\end{IEEEkeywords}

\section{Introduction}
Aerial robots have become practical tools for surveying, monitoring, and inspection because they can safely reach viewpoints that are difficult, costly, or hazardous for humans.
Civil \rev{Unmanned Aerial Systems (UAS)} activity and remote-pilot operations continue to grow, indicating broader operational adoption in public services and industry~\cite{faa2025forecast}.
Current deployments include logistics campuses, rail and port yards, construction corridors, and utility perimeter monitoring, where automated or semi-automated aerial observation is increasingly used~\cite{xu2023tunnel,xu2024drones360gimbal,liao2023birdsnest,yu2023transline}.

\begin{figure}[!t]
    \centering
    \begin{subfigure}[b]{\columnwidth}
        \centering
        \includegraphics[width=\textwidth]{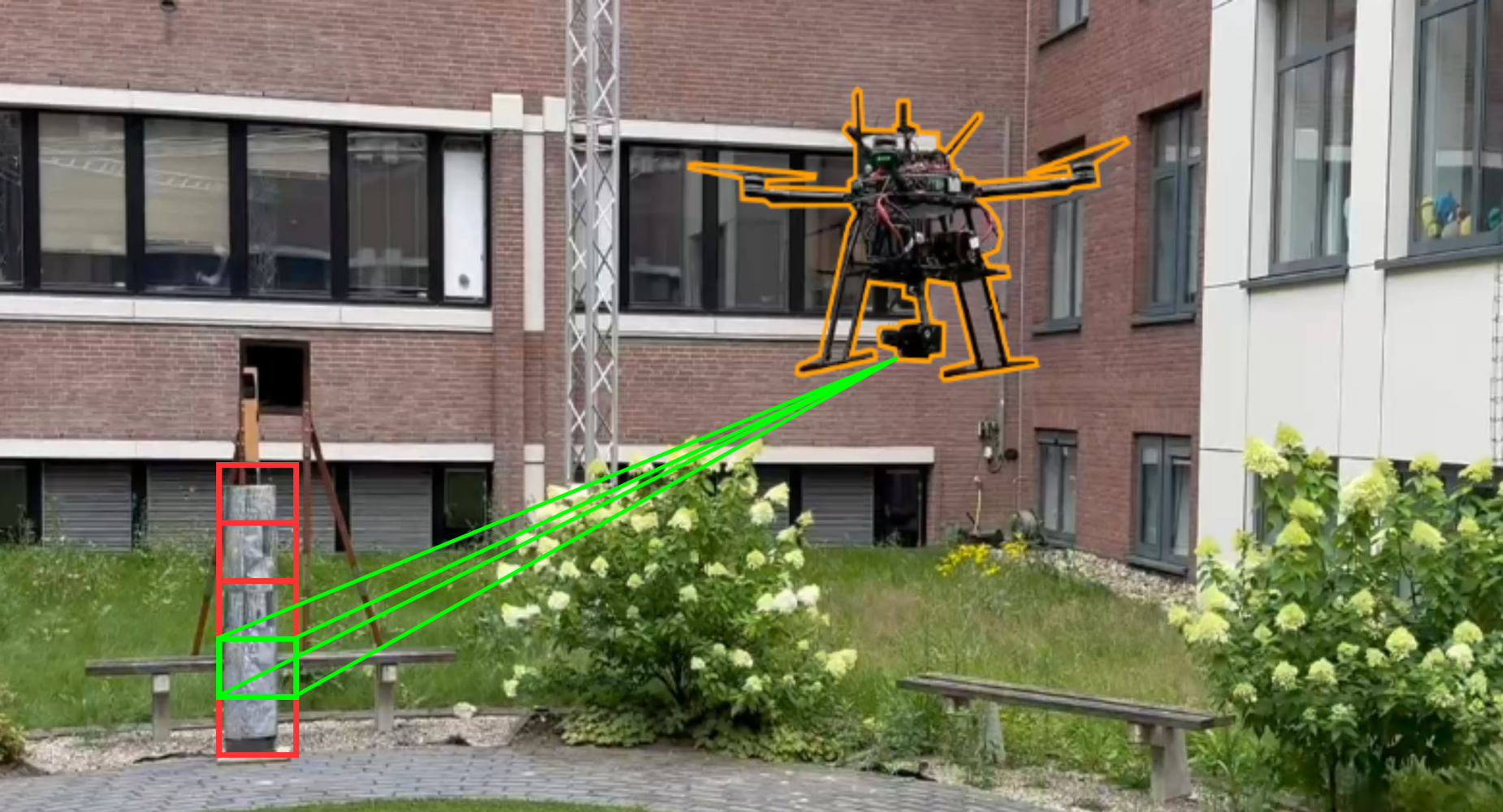}
        \caption{}
        \label{fig:cover-main}
    \end{subfigure}

    \vspace{1mm}

    \begin{subfigure}[b]{0.48\columnwidth}
        \centering
        \includegraphics[height=2.9cm,keepaspectratio]{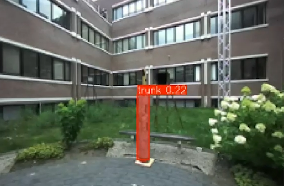}
        \caption{}
        \label{fig:cover-a}
    \end{subfigure}
    \hfill
    \begin{subfigure}[b]{0.48\columnwidth}
        \centering
        \includegraphics[height=2.9cm,keepaspectratio]{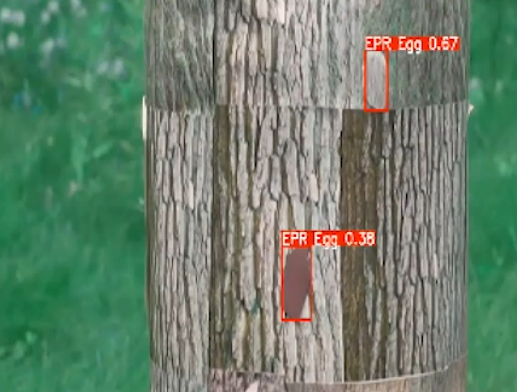}
        \caption{}
        \label{fig:cover-c}
    \end{subfigure}
    \caption{(a) \rev{Example use-case} of aerial micro-inspection for detecting \rev{small} oak processionary caterpillar egg clusters on a mockup tree trunk \rev{while keeping a safe distance}, (b) wide-view image from stereo camera \rev{(navigation camera)} required for target surface segmentation, partitioning the surface, gimbal planning, range estimation, and motion-compensation feedback, and (c) fine-detail inspection of a selected partition using the narrow-view camera \rev{(inspection camera)}.}
    \label{fig:cover}
\end{figure}

Beyond overview imaging, many applications require reliable inspection of small targets where aerial viewpoints are still beneficial \rev{or required}.
Examples include crack assessment in civil assets such as \rev{bridges and wind turbines}~\cite{humpe2020bridge360,xing2023uavcrack}, precision agriculture \rev{for crop disease and pest monitoring}~\cite{tsouros2019precisionagri}, security monitoring from distant airborne viewpoints~\cite{xu2024drones360gimbal}, and power-line corridor inspection for \rev{bird nests and hanging objects}~\cite{liao2023birdsnest,yu2023transline}.
Similar requirements appear in industrial surface inspection, where stable close-up imaging and reliable revisit are essential for defect detection \rev{in use cases such as photovoltaic module inspection and quality control in the automotive industry}~\cite{bergmann2019mvtecad,deitsch2019pvdefects,chang2019mobile}.
\rev{These are examples of what we refer to as \emph{Aerial Micro-Inspection}, where a small target requires high-detail imaging in a way that challenges the safety or dynamic constraints of an aerial robot.}

The need is therefore not only to fly and see, but to deliver reliable \rev{inspection} autonomy that reduces cost, improves repeatability, and limits personnel exposure.
\rev{This requires robust integration of perception, localization, planning, obstacle avoidance, trajectory following, and closed-loop target centering.
Many existing systems optimize only a subset of these components for specific scenarios, leaving generalization and modularity underexplored~\cite{nex2014uavreview,xu2024drones360gimbal,xu2023tunnel,bartczak2025orbit,simplicitio2024icuas_dams,yi2023proximitytracking,liu2020gimbaltracking,lin2021trackingstrategy}}.

Recent systems increasingly replace rigid body-fixed cameras with gimballed payloads~\cite{haneda2025viewpoint,wu2025ptz,xu2024drones360gimbal}.
This adds pan-tilt degrees of freedom and provides key benefits: sharper imagery through vibration isolation, pointing partially decoupled from quadrotor attitude, and reduced dependence on complex translational trajectories for coverage~\cite{haneda2025viewpoint,wu2025ptz}.
With zoom optics, narrow field-of-view (FOV) sensing can capture finer details from safer distances~\cite{wu2025ptz}.

\rev{We present \texttt{aerial\_micro\_inspection}, a generic pipeline compatible with the PX4 flight stack.
It adopts a dual-camera design~\cite{liu2020gimbaltracking,lin2021trackingstrategy,haneda2025viewpoint,wu2025ptz,xu2024drones360gimbal} that combines a wide-FOV stereo camera for on-site surface acquisition, partitioning, depth estimation, and drone motion compensation with a narrow-FOV gimballed camera for detailed inspection.
It requires no prior knowledge of the target geometry; instead, it integrates object detection/segmentation and stereo depth estimation for online surface geometry estimation.
}
The main contributions are:
\begin{enumerate}
    \item To the best of our knowledge, the first \rev{open-source} modular ROS 2-based pipeline explicitly focused on aerial micro-inspection.
    \item A parametric sweeping strategy to divide any inspection target into zoom-compliant partitions for coverage.
    \item A vision-based feedback loop that keeps gimbal pointing aligned with the intended partition under localization error and platform motion.
    \item An extrinsic calibration tool between a rigidly attached wide-view camera and a gimbal-mounted inspection camera.
\end{enumerate}

\rev{The rest of the paper is organized as follows: Section~\ref{sec:related_work} reviews related work in aerial inspection, dual-camera systems, and gimbal control. Section~\ref{sec:methodology} describes the kinematic model of the dual-camera system, the extrinsic calibration procedure, and the architecture of the proposed pipeline. Section~\ref{sec:implementation} presents simulation and real-world experiments in tree and greenhouse inspection scenarios, followed by a discussion in Section~\ref{sec:discussion}. Finally, Section~\ref{sec:conclusion} concludes the paper and discusses future work.}

\section{Related Work}
\label{sec:related_work}

A substantial portion of the aerial inspection literature assumes a rigidly mounted camera and frames inspection primarily as a flight-path planning problem under coverage, safety, and operational constraints. 
For example, ORBIT~\cite{bartczak2025orbit} provides an open-source toolkit for bridge inspection mission planning, generating coordinated waypoint routes for overview and underdeck surveys under realistic constraints. 
In confined or GNSS-challenged environments, many systems focus on autonomy stacks that tightly integrate task logic with localization, mapping, and collision-free navigation. 
Xu \emph{et al.}~\cite{xu2023tunnel} propose a vision-based autonomous unmanned aerial vehicle (UAV) inspection framework for unknown tunnel construction sites with dynamic obstacles, using hierarchical decision-making and planning, a dynamic mapping module, and post-flight reconstruction. 
Following similar themes, inspection is often coupled to photogrammetry/mapping mission design and coverage-style trajectories.
For instance, Simplicio and Pereira~\cite{simplicitio2024icuas_dams} design photogrammetry-based mission planning for dam mapping, including resolution/overlap-driven flight proximity and coverage rows. 
While these approaches enable repeatable inspection and robust navigation, their primary target is commonly path planning, localization/mapping, and safety/avoidance, which can lead to longer flight paths, increased operational risk when close proximity is required for high resolution, and, in many cases, dependence on prior structure information or maps for planning and safety margins~\cite{bartczak2025orbit,xu2023tunnel,simplicitio2024icuas_dams}.

Another line of work addresses inspection by generating viewpoints (or camera configurations) to guarantee coverage and spatial resolution. 
Haneda \emph{et al.}~\cite{haneda2025viewpoint} propose a viewpoint generation algorithm for gimbal-mounted robots to inspect civil engineering structures while leveraging gimbal degrees of freedom to reduce base motion. 
However, such planning-centric formulations typically do not explicitly account for localization inaccuracies and constant multirotor wobble, and therefore do not ensure that the gimbal is aligned with the intended regions at the correct time.
Wu \emph{et al.}~\cite{wu2025ptz} provide a related configuration-planning approach for pan-tilt-zoom (PTZ) camera-equipped robots, explicitly planning PTZ configurations to meet required spatial resolution. 
They divide the target surface into patches and plan PTZ configurations to cover each patch, but they do not consider the dynamics of the system or the potential for target drift, which can lead to misalignment between the planned configurations and the actual inspection targets, especially if the robot is in flight.

Inspection payloads have advanced to include multiple cameras with different focal lengths mounted on the same gimbal, often configured as \textit{Peripheral-Central} or \textit{Foveated} setups~\cite{ude2006foveated, zhang2024foveacam++}. 
Inspired by human vision, these systems combine a wide peripheral view for situational awareness with a narrow view for detailed analysis, and can be useful for aerial micro-inspection; however, many rely on specialized gimbal payloads that increase cost and system complexity. 
Related UAV designs place the peripheral camera on the vehicle body while keeping the central camera on the gimbal~\cite{kang2021development}, which broadens hardware compatibility. 
This design allows the peripheral camera to support state estimation, obstacle avoidance, and navigation tasks while being used to guide the gimbal. 
Similarly, Xu \emph{et al.}~\cite{xu2024drones360gimbal} combine a panoramic wide-view sensor with a gimbal camera for detection and tracking, using the wide view for initial discovery and the gimbal payload for higher-definition follow-up. These studies motivate dual-camera perception architectures, but they typically target surveillance or tracking rather than guaranteeing fine-detailed surface coverage and reliable micro-inspection under aerial disturbances.

\section{Methodology}
\label{sec:methodology}
\rev{Section~\ref{sec:kinematics} introduces the kinematic model of the dual-camera system, defining the key variables, parameters, and assumptions.
Section~\ref{sec:extrinsic_calibration} describes the extrinsic calibration procedure between the navigation camera and the gimbal base, a necessary prerequisite for the pipeline.
Finally, Section~\ref{sec:pipeline} presents the pipeline architecture and data flow between its components, including the partitioning algorithm.}

\subsection{Kinematics Model}
\label{sec:kinematics}
\begin{figure}[tb]
    \centering
    \includegraphics[width=0.5\textwidth]{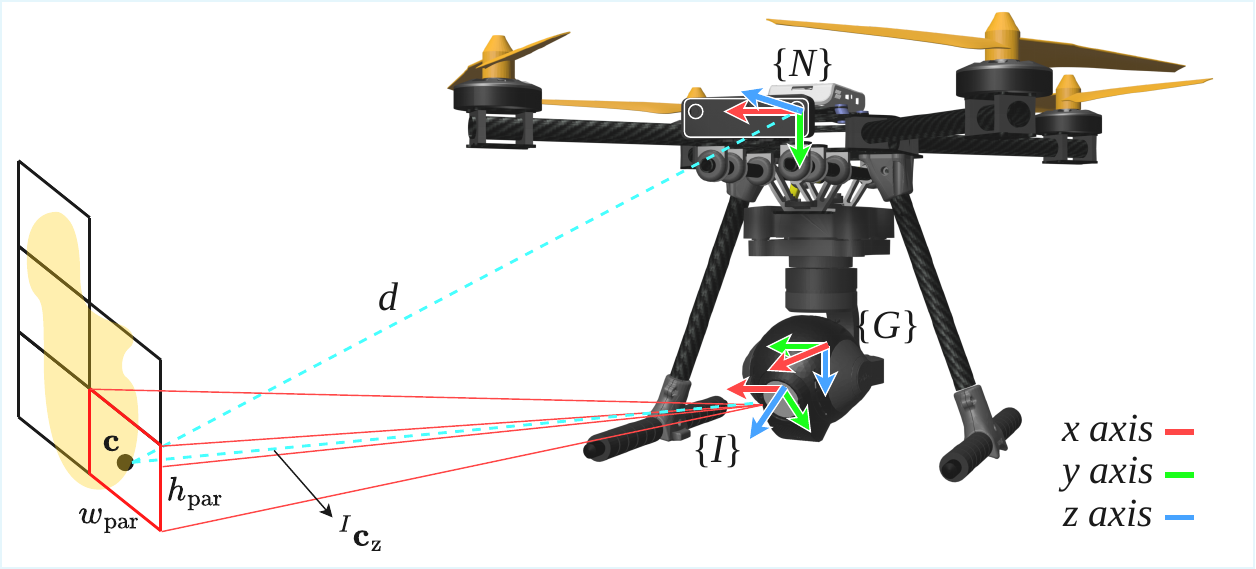}
    \caption{Dual-camera inspection kinematic model. The yellow blob represents the target surface; the surface is partitioned and each partition is inspected sequentially by aligning the gimbal to the corresponding direction.}
    \label{fig:kinematic_model}
\end{figure}

\rev{Throughout this manuscript, we use a notation that describes a point $\mathbf{a}$ expressed in a coordinate frame $\{X\}$ using ${}^{X}\mathbf{a}$.
For a homogeneous transformation matrix that converts a point from coordinate system $\{Y\}$ to $\{X\}$, we use ${}^{X}\mathbf{T}_{Y}$. 
Each ${}^{X}\mathbf{T}_{Y}$ can be decomposed into a rotation matrix ${}^{X}\mathbf{R}_{Y}$ and a translation vector ${}^{X}\mathbf{t}_{Y}$ such that
\begin{equation}
{}^{X}\mathbf{T}_{Y} = \begin{bmatrix}
    {}^{X}\mathbf{R}_{Y} & {}^{X}\mathbf{t}_{Y} \\
    \mathbf{0} & 1
\end{bmatrix}.
\end{equation}}

We consider an aerial robot equipped with a stereo wide-angle \textit{navigation camera} and a gimballed narrow-FOV \textit{inspection camera}.
As described in Fig.~\ref{fig:kinematic_model}, the main reference coordinates are the navigation camera's optical frame $\{N\}$, gimbal base frame $\{G\}$, gimbal instantaneous frame $\{G_t\}$, inspection camera frame $\{C\}$, and inspection camera's optical frame $\{I\}$.
The gimbal base frame is located at the gimbal's center of rotation and is aligned with the zero pan-tilt configuration.
The gimbal instantaneous frame is the result of rotating the gimbal base frame by the current pan and tilt angles.

\rev{We need a kinematic model $f$ that relates a pixel in navigation camera's image plane ${}^{N}\mathbf{u}$ to a pixel in the inspection camera's image plane ${}^{I}\mathbf{u}$, which is necessary for pointing the gimbal to the correct location on the target surface. 
This non-linear mapping can be described as
}
\begin{equation}
    {}^{I}\mathbf{u} = f\!\left({}^{N}\mathbf{u}, d, {}^{G_t}\mathbf{T}_{G}\,;\,\boldsymbol{\theta}\right),
\end{equation}
\rev{where $d$ is the depth value corresponding to ${}^{N}\mathbf{u}$ pixel and ${}^{G_t}\mathbf{T}_{G}$ is the transformation caused by changing the gimbal's orientation.
Both $d$ and ${}^{G_t}\mathbf{T}_{G}$ are variables.
The $f$ also depends on a set of parameters $\boldsymbol{\theta}$ which are}
\begin{equation}
    \boldsymbol{\theta} = \left\{\mathbf{K}_\mathrm{nav}, \mathbf{K}    _\mathrm{ins}, {}^{G}\mathbf{T}_{N}, {}^{C}\mathbf{T}_{G_t}, {}^{I}\mathbf{T}_{C}\right\}.
\end{equation}
\rev{where $\mathbf{K}_\mathrm{nav}$ and $\mathbf{K}_\mathrm{ins}$ are the intrinsic parameters of the navigation and inspection cameras, respectively, ${}^{G}\mathbf{T}_{N}$ is the extrinsic transform between the navigation camera optical frame and the gimbal base frame, ${}^{C}\mathbf{T}_{G_t}$ is the extrinsic transform caused by mechanical misalignment of the inspection camera on the gimbal mount, and ${}^{I}\mathbf{T}_{C}$ is the transform between inspection camera and its optical frame.
The full kinematic chain in $f$ is explained in Appendix~\ref{app:kinematics}.} 

\begin{figure}[!t]
    \centering
    \begin{subfigure}[b]{0.45\columnwidth}
        \centering
        \includegraphics[height=2.7cm,keepaspectratio]{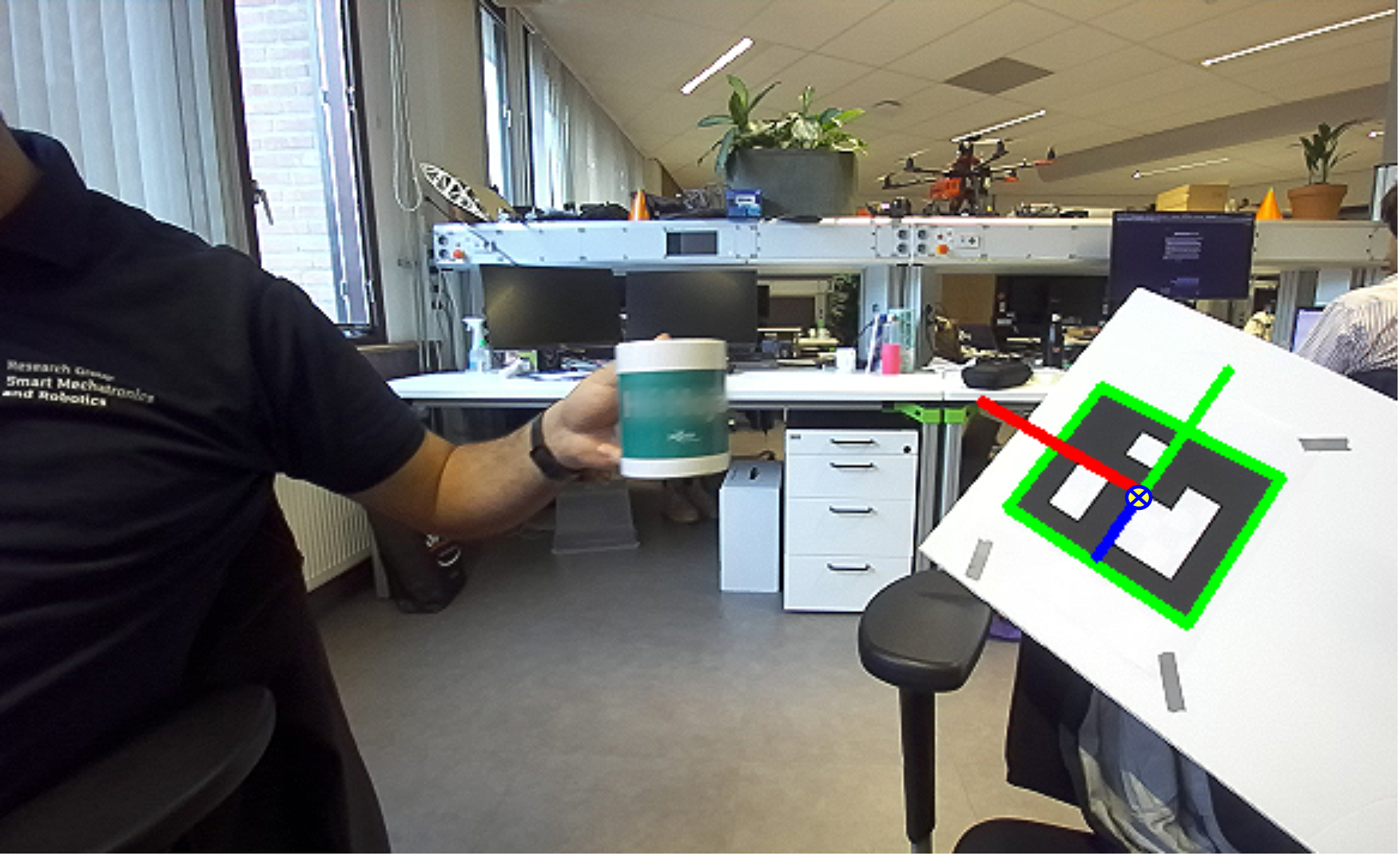}
        \caption{}
        \label{fig:calib-nav}
    \end{subfigure}
    \hfill
    \begin{subfigure}[b]{0.45\columnwidth}
        \centering
        \includegraphics[height=2.7cm,keepaspectratio]{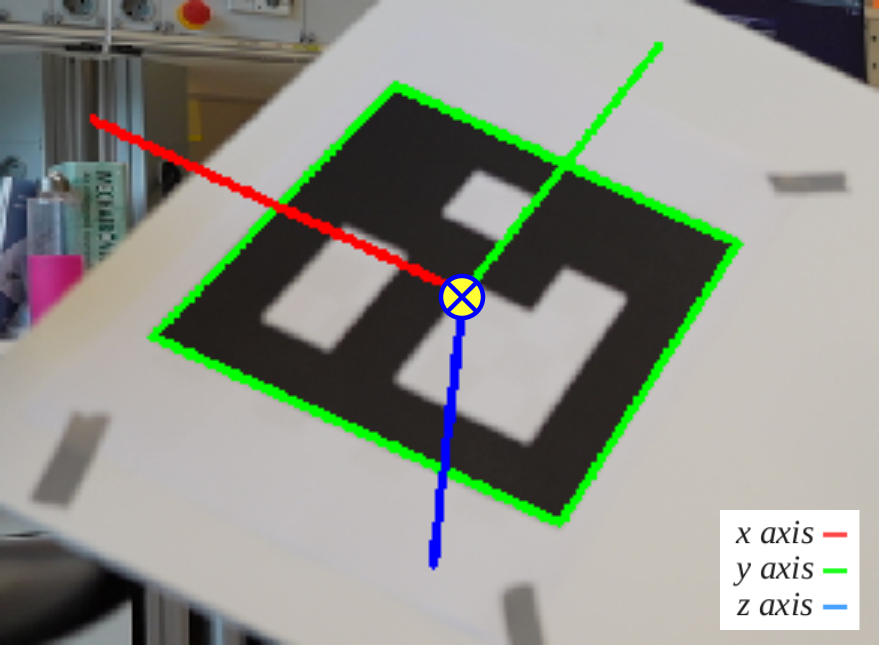}
        \caption{}
        \label{fig:calib-inspection}
    \end{subfigure}
    \caption{Calibration using an ArUco marker as a target. \rev{In} (a) the marker is observed by the navigation camera, and \rev{in} (b) the marker is observed by the inspection camera.}
    \label{fig:calib}
\end{figure}

\begin{figure*}[!t]
    \centering
    \includegraphics[width=\textwidth]{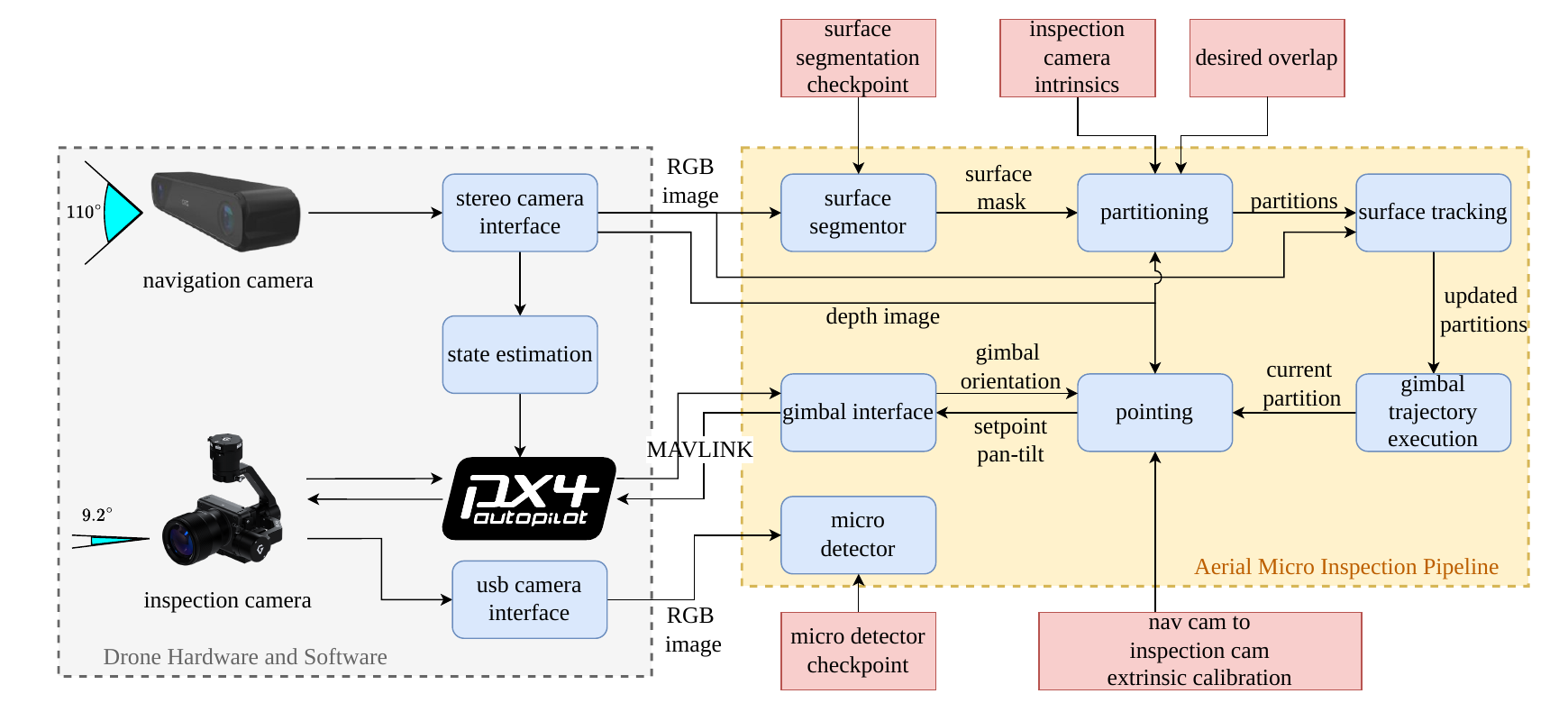}
    \caption{Schematic overview of the proposed aerial micro-inspection pipeline. In this diagram, the processes necessary in our modular pipeline are grouped and separated in the yellow box. The gray box represents components and processes related to a relatively interchangeable aerial robot design. Finally, the red blocks introduce parameters to adjust the pipeline to a different use case.}
    \label{fig:pipeline}
\end{figure*}

\rev{In our dual camera formulation, we make the assumption that the inspection target is within a range where stereo depth estimation from the navigation camera can estimate the $d$ variable. 
We also assume that the gimbal rotates perfectly around its center of rotation (the pan-tilt axes are intersecting and orthogonal), meaning that the variable ${}^{G_t}\mathbf{T}_{G}$ is a pure rotation transform that can be directly obtained from the gimbal's reported pan and tilt angles.
To obtain the parameter ${}^{G}\mathbf{T}_{N}$ we introduce an extrinsic calibration tool in the next subsection and we further assume that the mechanical misalignment of the inspection camera on the gimbal mount is negligible, meaning that ${}^{C}\mathbf{T}_{G_t}$ is an identity transform and ${}^{I}\mathbf{T}_{C}$ is only a re-axis transform without translation because the inspection camera's optical frame is aligned with its body frame.}


\subsection{Extrinsic Calibration of the Dual-Camera System}
\label{sec:extrinsic_calibration}
The problem of extrinsic calibration between two or more rigidly attached cameras is well studied with many open-source tools such as \texttt{kalibr}~\cite{furgale2013unified}.
Cases where both cameras have free degrees of freedom for pan-tilt-zoom (gimbal for pan-tilt) are also investigated~\cite{mao2022general}.
We adopt a similar target-based calibration approach for the case where one camera is rigidly attached to the drone and the other has pan-tilt degrees of freedom in order to obtain the extrinsic parameters ${}^{G}\mathbf{R}_{N}$ and ${}^{G}\mathbf{t}_{N}$.

As depicted in Fig.~\ref{fig:calib}, we use an ArUco marker as the calibration target, which is observed by both cameras.
Assuming that the intrinsic parameters of each camera ($\mathbf{K}_\mathrm{nav}$ and $\mathbf{K}_\mathrm{ins}$) are acquired by standard tools such as \texttt{ros\_camera\_calibration}~\cite{roscameracalibration}, and having the dimensions of the marker, we can get the 3D pose of the marker in each camera's frame by using the \rev{\texttt{estimatePoseSingleMarkers}} function from OpenCV.
We denote the pose of the marker in the navigation camera frame at time $k$ as ${}^{N}\mathbf{T}_{M}(k)$ and the pose of the marker in the inspection camera frame as ${}^{I}\mathbf{T}_{M}(k)$.
The transform from the inspection camera frame to the navigation camera frame will be
\begin{equation}
{}^{N}\mathbf{T}_{I}(k) = {}^{N}\mathbf{T}_{M}(k)\;{}^{I}\mathbf{T}_{M}(k)^{-1}.
\end{equation}
The transform from the inspection camera as described in Appendix~\ref{app:kinematics} is
\begin{equation}
{}^{N}\mathbf{T}_{I}(k) = {}^{N}\mathbf{T}_{G}(k)\;{}^{G}\mathbf{T}_{G_t}(k)\;{}^{G_t}\mathbf{T}_{I}.
\end{equation}  
Solving for ${}^{N}\mathbf{T}_{G}$, we have
\begin{equation}
{}^{N}\mathbf{T}_{G}(k) = {}^{N}\mathbf{T}_{M}(k)\;{}^{I}\mathbf{T}_{M}(k)^{-1}\;{}^{G_t}\mathbf{T}_{I}^{-1}\;{}^{G}\mathbf{T}_{G_t}(k)^{-1}.
\end{equation}

Assuming limited variation in ${}^{N}\mathbf{T}_{G}(k)$ (which requires a stationary target and fixed gimbal orientation in an unsynchronized setup), we average the translational components to obtain ${}^{N}\mathbf{t}_{G}$ and use \rev{Singular Value Decomposition (SVD)} to average the rotational components to obtain ${}^{N}\mathbf{R}_{G}$.

\subsection{Aerial Micro Inspection Pipeline}
\label{sec:pipeline}

Figure~\ref{fig:pipeline} provides a schematic overview of the proposed aerial micro-inspection pipeline.
This subsection summarizes its main components and data flow.

The navigation camera (also referred to as the peripheral camera) provides a wide field of view in front of the drone.
Many aerial inspection \rev{methods} place both \rev{navigation and inspection cameras} on the same gimbal~\cite{ude2006foveated, zhang2024foveacam++}.
Another approach is to mount the \rev{navigation} camera on the drone body and the \rev{inspection} camera on the gimbal~\cite{kang2021development,xu2024drones360gimbal}, which is the design we consider in this work.
This choice allows the \rev{navigation} camera to also support state estimation and navigation modules (Fig.~\ref{fig:pipeline}).

The RGB image is processed by the \textit{Surface Segmentor}, \rev{responsible for detecting the target surface}, currently implemented as a wrapper around YOLO-based detection/segmentation models~\cite{yolov8_ultralytics}.
The corresponding model checkpoint is provided as a runtime parameter.
Its output is the segmented target surface, which is passed to the partitioning module.

In the \textit{Partitioning} block, we divide the target surface into smaller regions of interest, each approximately matching the projected field of view of the inspection camera. 
Therefore, we calculate the projected size of the inspection camera's field of view in the navigation camera's image plane using the intrinsic parameters of both cameras:
\begin{equation}
    (w_{\mathrm{par}},h_{\mathrm{par}}) = \left(\frac{f_{x_{\mathrm{nav}}}}{f_{x_{\mathrm{ins}}}}W_\mathrm{nav}, \frac{f_{y_{\mathrm{nav}}}}{f_{y_{\mathrm{ins}}}}H_\mathrm{nav}\right),
\end{equation}
where $f_{x_{\mathrm{nav}}}$ and $f_{y_{\mathrm{nav}}}$ are the focal lengths of the navigation camera, $f_{x_{\mathrm{ins}}}$ and $f_{y_{\mathrm{ins}}}$ are the focal lengths of the inspection camera, and $W_\mathrm{nav}$ and $H_\mathrm{nav}$ are the width and height of the navigation camera's image.
\rev{The calculated $(w_{\mathrm{par}},h_{\mathrm{par}})$ is also visualized in Fig.~\ref{fig:kinematic_model}.}

\begin{algorithm}[t]
\caption{Partitioning of the Segmented Surface in the Image Plane}
\label{alg:sweep_partitioning}
\begin{algorithmic}[1]
\REQUIRE Binary mask $M \in \{0,1\}^{H \times W}$, window size $(w_{\mathrm{par}},h_{\mathrm{par}})$, overlap $o \in [0,1)$, occupancy threshold $\tau \in (0,1]$
\ENSURE Ordered list of accepted windows $\mathcal{B}$
\STATE $\mathcal{B} \leftarrow [\ ]$, $A \leftarrow w_{\mathrm{par}} \cdot h_{\mathrm{par}}$
\STATE $(\mathcal{Y},\mathcal{X}) \leftarrow \{(y,x)\mid M[y,x]=1\}$
\STATE $x_{\min} \leftarrow \min(\mathcal{X}),\; x_{\max} \leftarrow \max(\mathcal{X})$
\STATE $y_{\min} \leftarrow \min(\mathcal{Y}),\; y_{\max} \leftarrow \max(\mathcal{Y})$
\STATE $s_x \leftarrow \lfloor w_{\mathrm{par}}(1-o)\rfloor$ 
\STATE $s_y \leftarrow \lfloor h_{\mathrm{par}}(1-o)\rfloor$
\STATE Build $X_0$ from $x=x_{\min}:s_x:x_{\max}$ 
\STATE Build $Y_0$ from $y=y_{\min}:s_y:y_{\max}$ then reverse order
\FOR{$y_0 \in Y_0$}
    \STATE $y_1 \leftarrow y_0 + h_{\mathrm{par}}$
    \FOR{$x_0 \in X_0$}
        \STATE $x_1 \leftarrow x_0 + w_{\mathrm{par}}$
        \STATE $c \leftarrow \sum M[y_0:y_1,\,x_0:x_1]$
        \IF{$c \ge \tau A$}
            \STATE Append $(x_0,y_0,x_1,y_1)$ to $\mathcal{B}$
        \ENDIF
    \ENDFOR
\ENDFOR
\STATE \textbf{return} $\mathcal{B}$
\end{algorithmic}
\end{algorithm}

Algorithm~\ref{alg:sweep_partitioning} formalizes partitioning of the target surface in image space so the inspection camera can sweep it.
The method first finds the segmented mask $M \in \{0,1\}^{H \times W}$ extents, computes scan strides with some \rev{user-defined} overlap $o \in [0,1)$, and then sweeps bottom-to-top and left-to-right. 
A candidate window is accepted only when the segmented-pixel occupancy inside that window exceeds a threshold $\tau \in (0,1]$.

Partitioning is not performed every frame; it is triggered at the beginning of a sweep.
The resulting partitions are then visited sequentially by pointing the gimbal to each partition center.
Because the drone's motion \rev{during the time that the gimbal is visiting all partitions} can shift the apparent surface location, the pipeline includes \rev{the \textit{Surface Tracking} module}.
The tracker is initialized with the surface bounding box whenever partitioning is updated.
For each frame, it returns $(du,dv)$ pixel offsets that are applied to all partition centers.
We used the Median Flow tracker~\cite{kalal2010forward} from OpenCV.

\begin{figure}[!t]
    \centering
    \includegraphics[width=0.48\textwidth]{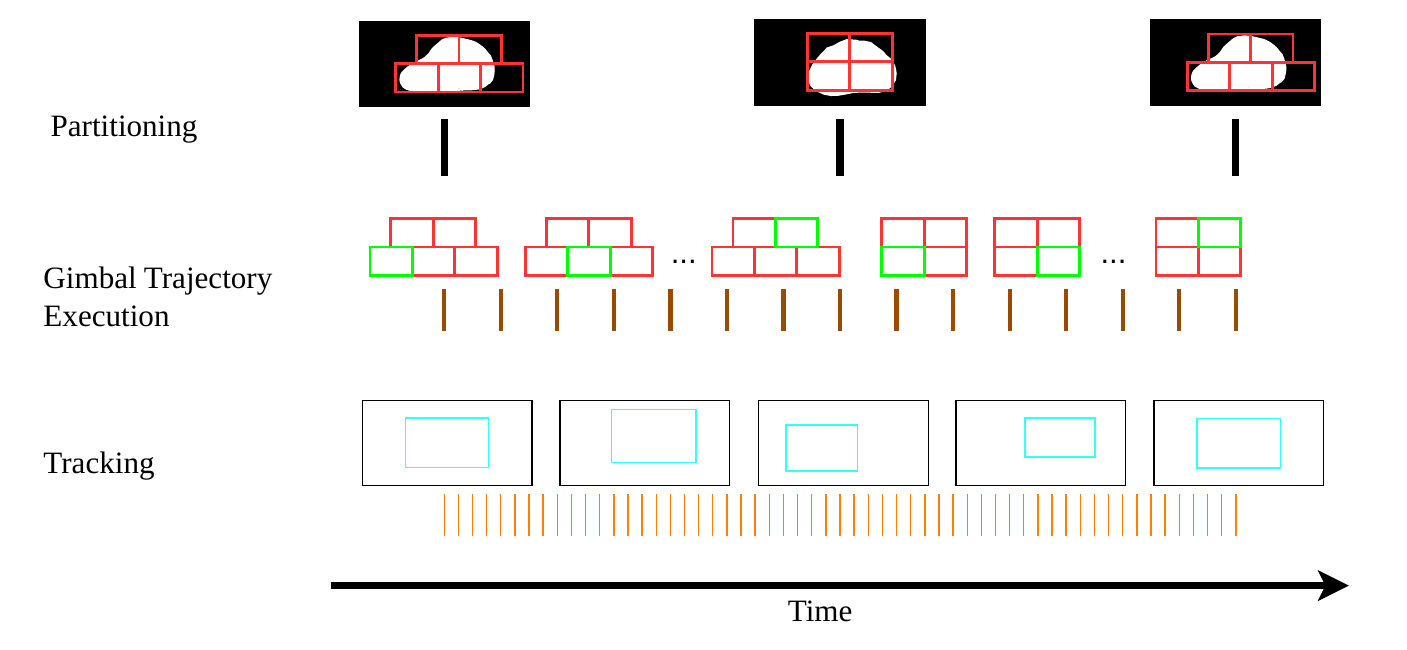}
    \caption{Relative time frequency of partitioning, tracking, and gimbal trajectory execution in the proposed pipeline. Each vertical is an instance of time that the respective task is executed. \rev{The partitioning task runs at lowest frequency, such that the entire partitions should be visited by inspection camera before the next partitioning. Gimbal trajectory execution happens at intervals configurable by the user (e.g., every 2 seconds); the gimbal locks on the current partition for some time and moves to next partition. Tracking happens at camera frame-rate, making sure the location of all partitions are updated in real-time to compensate for drone motion.}}
    \label{fig:freqs}
\end{figure}

Figure~\ref{fig:freqs} shows that partitioning runs at a lower frequency than tracking and gimbal execution.
The exact partition set may vary frame to frame because of segmentation jitter and drone motion.
Therefore, a sweep should complete before re-partitioning; during a sweep, partition locations are updated using tracker offsets.

\begin{figure*}[t]
    \centering
    \begin{minipage}[t]{0.24\textwidth}
        \centering
        \begin{subfigure}[t]{\linewidth}
            \centering
            \includegraphics[width=\linewidth,height=2.4cm]{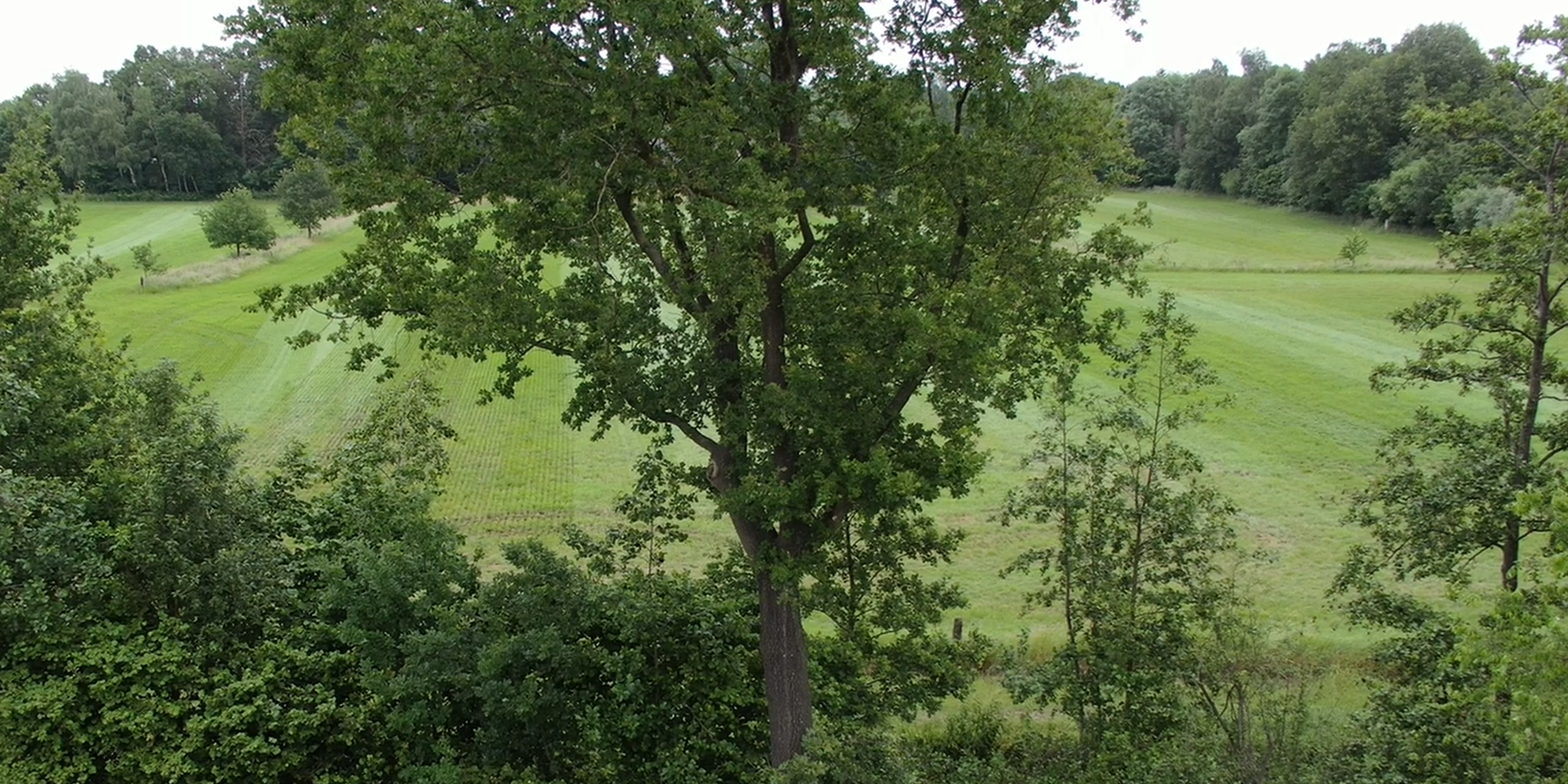}
            \caption{}
            \label{fig:impl-a}
        \end{subfigure}

        \vspace{1mm}
        \begin{subfigure}[t]{\linewidth}
            \centering
            \includegraphics[width=\linewidth,height=2.4cm]{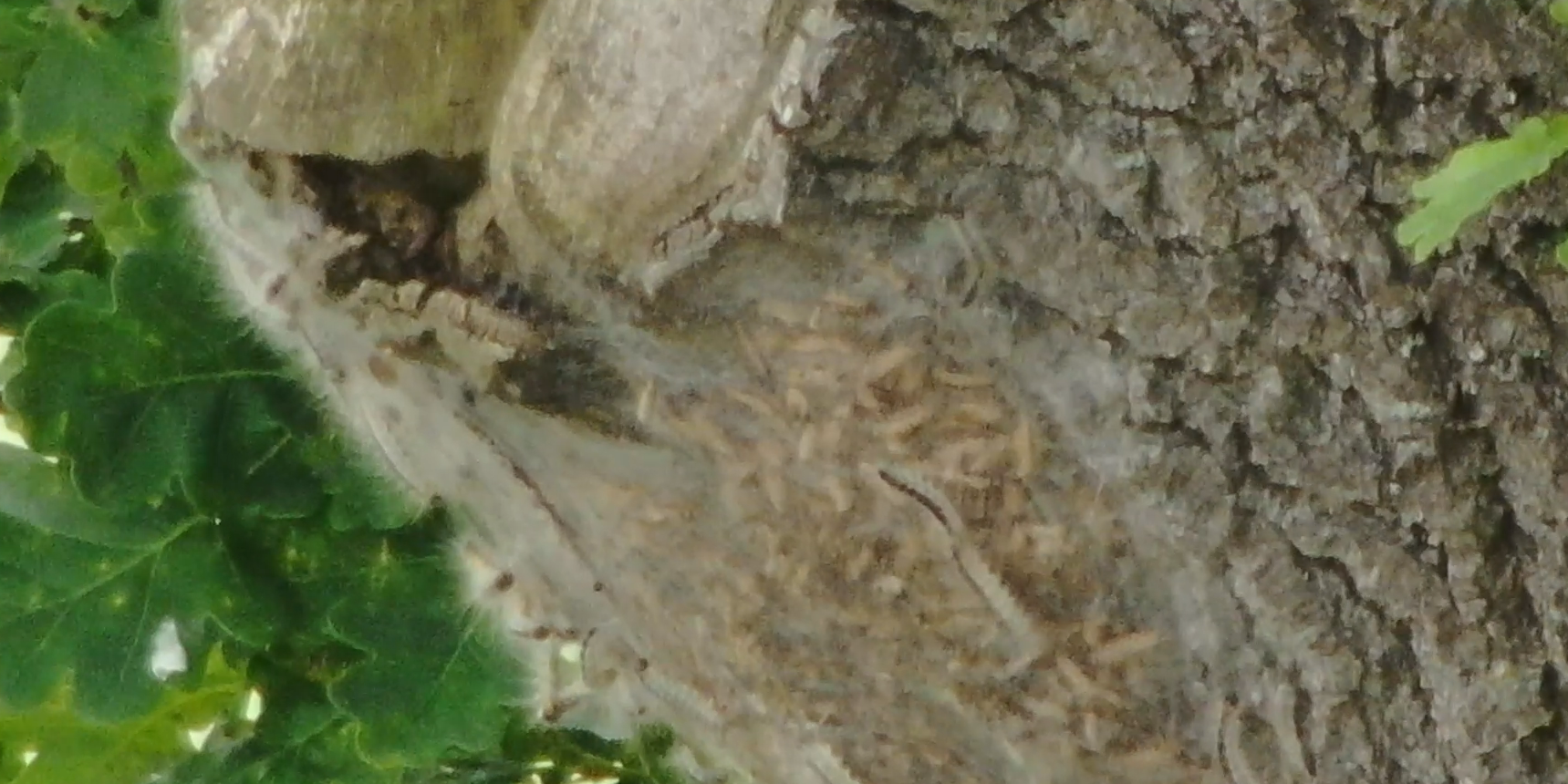}
            \caption{}
            \label{fig:impl-b}
        \end{subfigure}
    \end{minipage}
    \hfill
    \begin{minipage}[t]{0.24\textwidth}
        \centering
        \begin{subfigure}[t]{\linewidth}
            \centering
            \includegraphics[width=\linewidth,height=2.4cm]{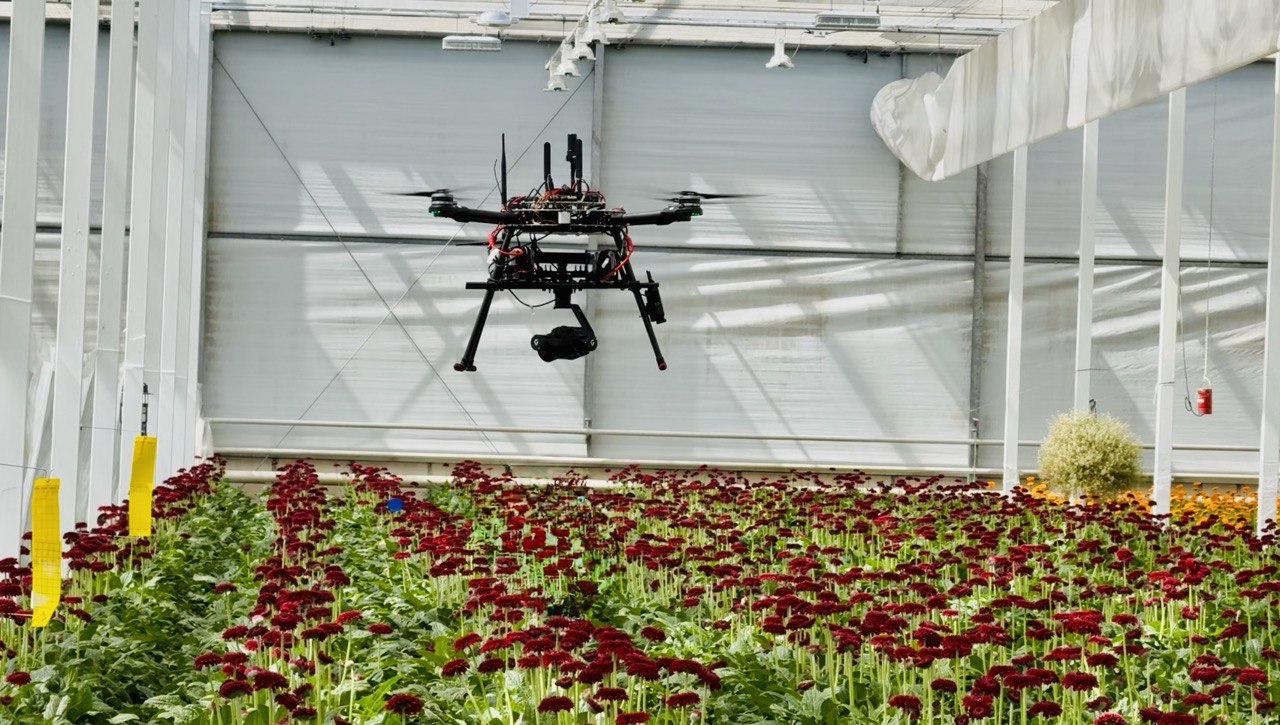}
            \caption{}
            \label{fig:impl-c}
        \end{subfigure}

        \vspace{1mm}
        \begin{subfigure}[t]{\linewidth}
            \centering
            \includegraphics[width=\linewidth,height=2.4cm]{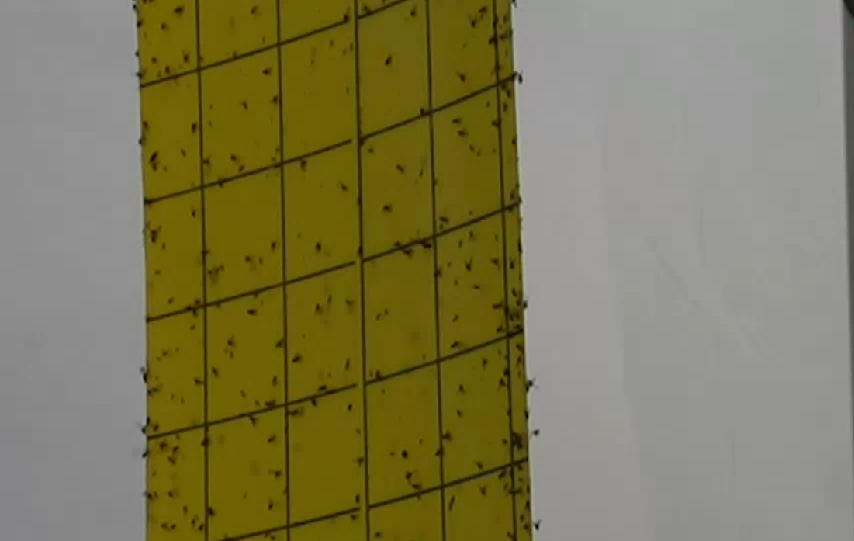}
            \caption{}
            \label{fig:impl-d}
        \end{subfigure}
    \end{minipage}
    \hfill
    \begin{minipage}[t]{0.24\textwidth}
        \centering
        \begin{subfigure}[t]{\linewidth}
            \centering
            \includegraphics[width=\linewidth,height=2.4cm]{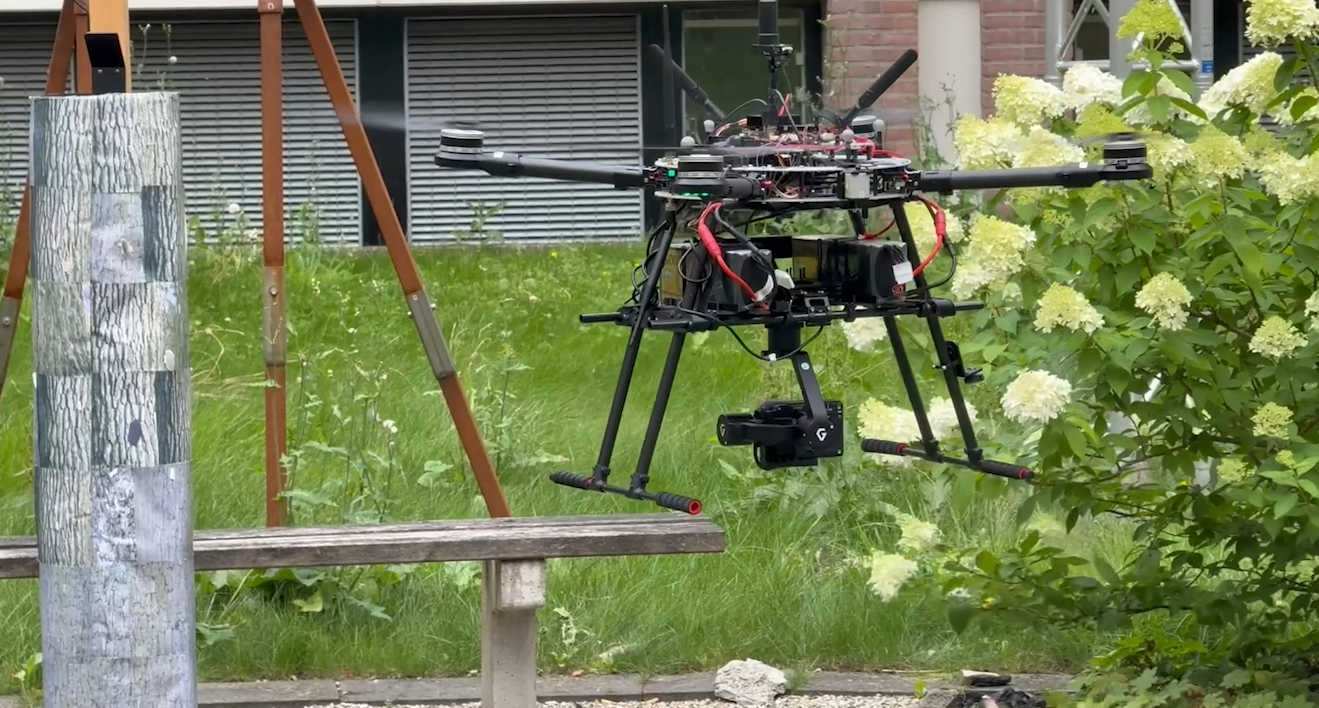}
            \caption{}
            \label{fig:impl-e}
        \end{subfigure}

        \vspace{1mm}
        \begin{subfigure}[t]{\linewidth}
            \centering
            \includegraphics[width=\linewidth,height=2.4cm]{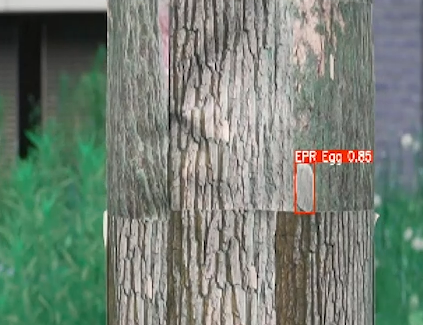}
            \caption{}
            \label{fig:impl-f}
        \end{subfigure}
    \end{minipage}
    \hfill
    \begin{minipage}[t]{0.24\textwidth}
        \centering
        \begin{subfigure}[t]{\linewidth}
            \centering
            \includegraphics[width=\linewidth,height=2.4cm]{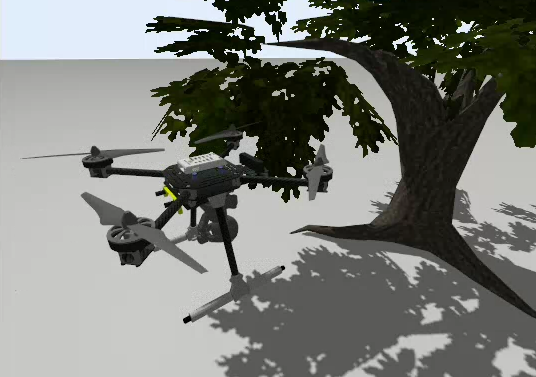}
            \caption{}
            \label{fig:impl-g}
        \end{subfigure}

        \vspace{1mm}
        \begin{subfigure}[t]{\linewidth}
            \centering
            \includegraphics[width=\linewidth,height=2.4cm]{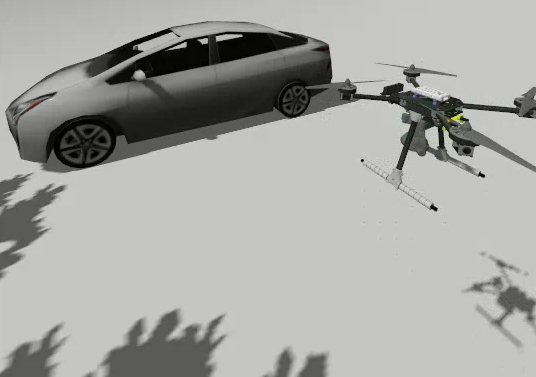}
            \caption{}
            \label{fig:impl-h}
        \end{subfigure}
    \end{minipage}
    \caption{Example scenes and deployments. (a,b) Wide and close-up views of an oak tree affected by caterpillar nests. (c,d) Greenhouse deployment with sticky-trap inspection. (e,f) Mockup-tree experiments with artificial caterpillar eggs and corresponding inspection views. (g,h) Gazebo simulation environments.}
    \label{fig:impl}
\end{figure*}

\begin{figure*}[!t]
    \centering
    \begin{subfigure}[b]{0.19\textwidth}
        \centering
        \includegraphics[width=\linewidth,height=2.5cm]{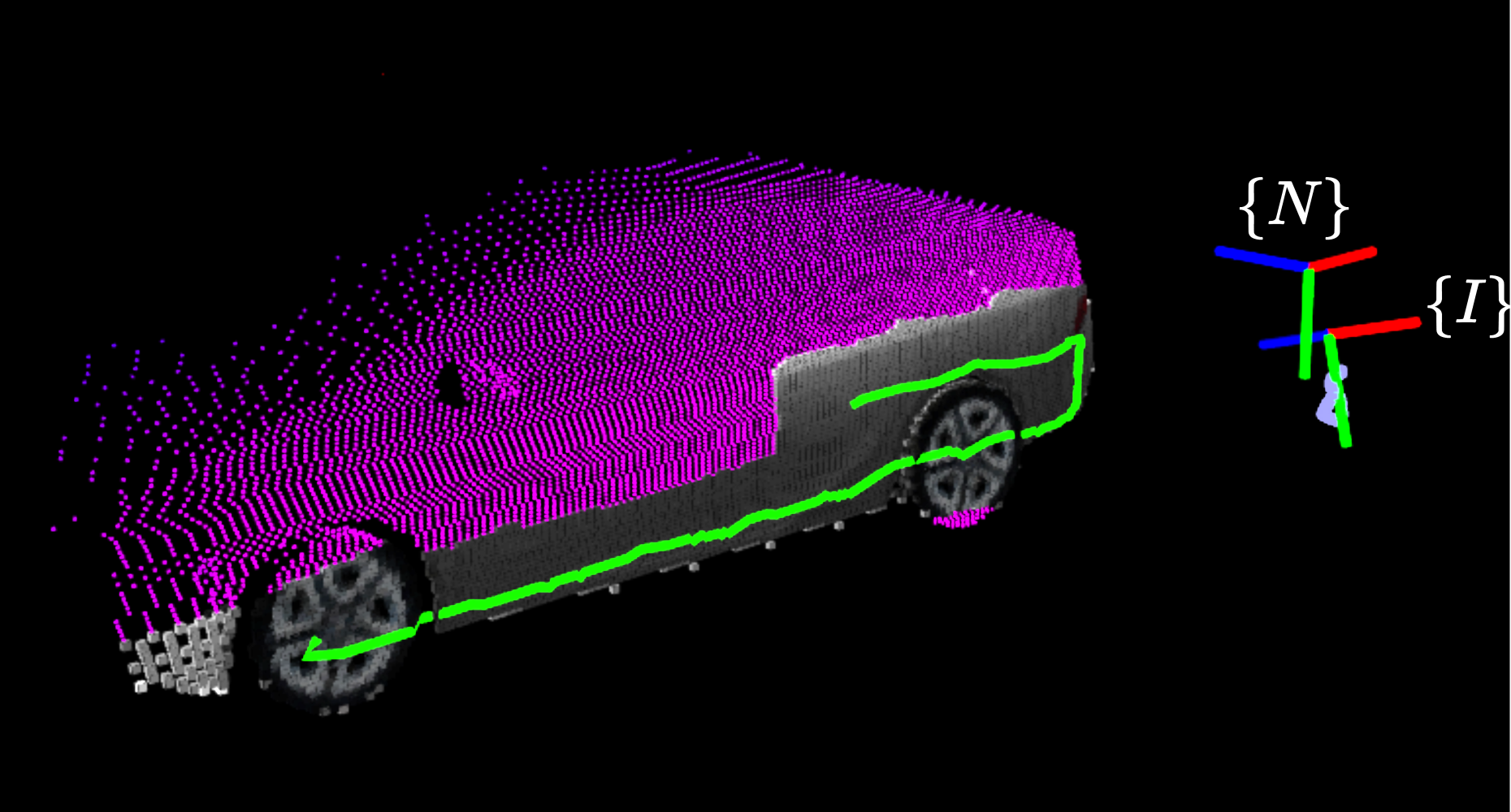}
        \caption{}
        \label{fig:coveage_vis}
    \end{subfigure}
    \hfill
    \begin{subfigure}[b]{0.19\textwidth}
        \centering
        \includegraphics[width=\linewidth,height=2.5cm]{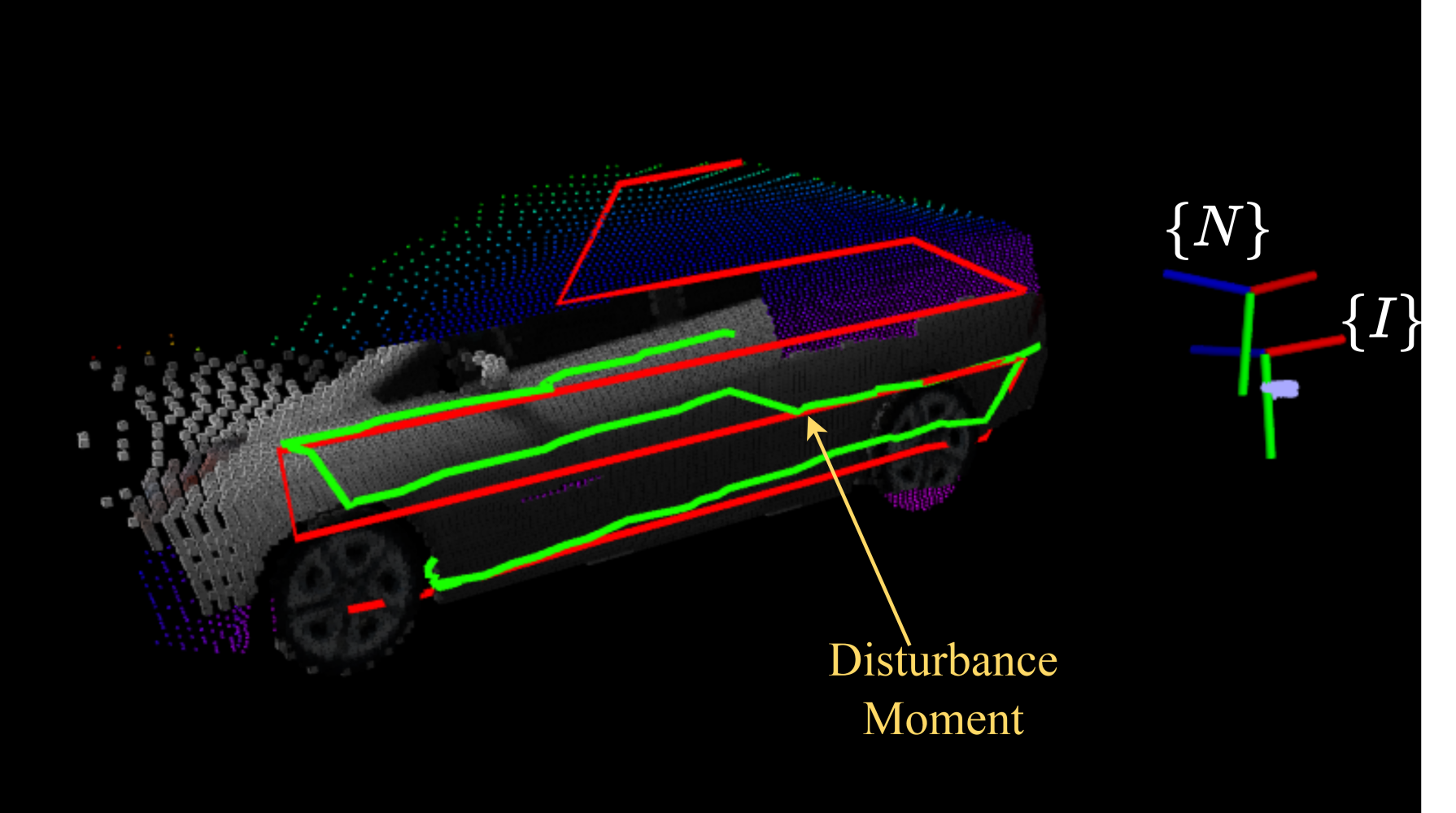}
        \caption{}
        \label{fig:disturbance_ff}
    \end{subfigure}
    \hfill
    \begin{subfigure}[b]{0.19\textwidth}
        \centering
        \includegraphics[width=\linewidth,height=2.5cm]{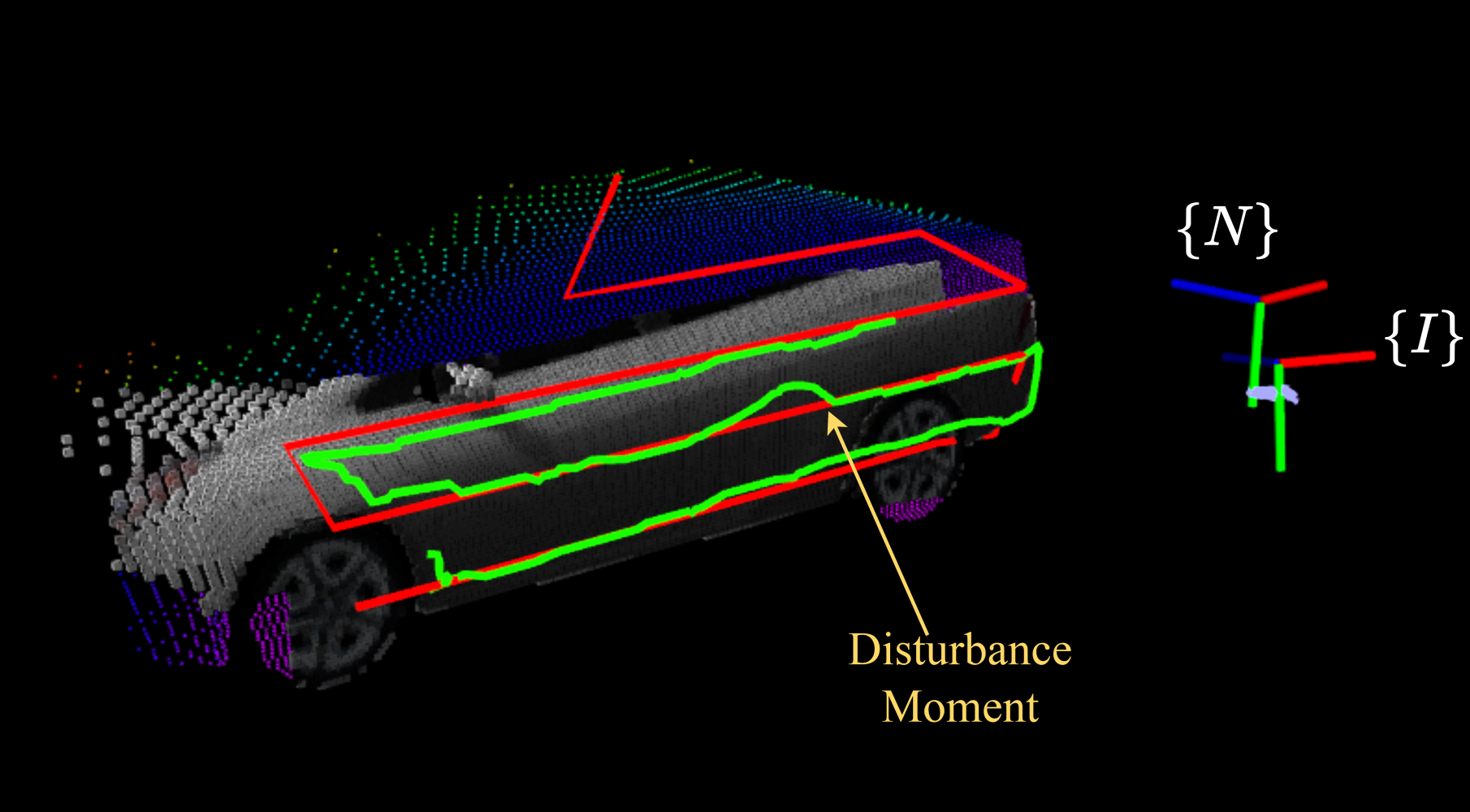}
        \caption{}
        \label{fig:disturbance_fb}
    \end{subfigure}
    \hfill
    \begin{subfigure}[b]{0.19\textwidth}
        \centering
        \includegraphics[width=\linewidth,height=2.5cm,trim=6 10 6 10,clip]{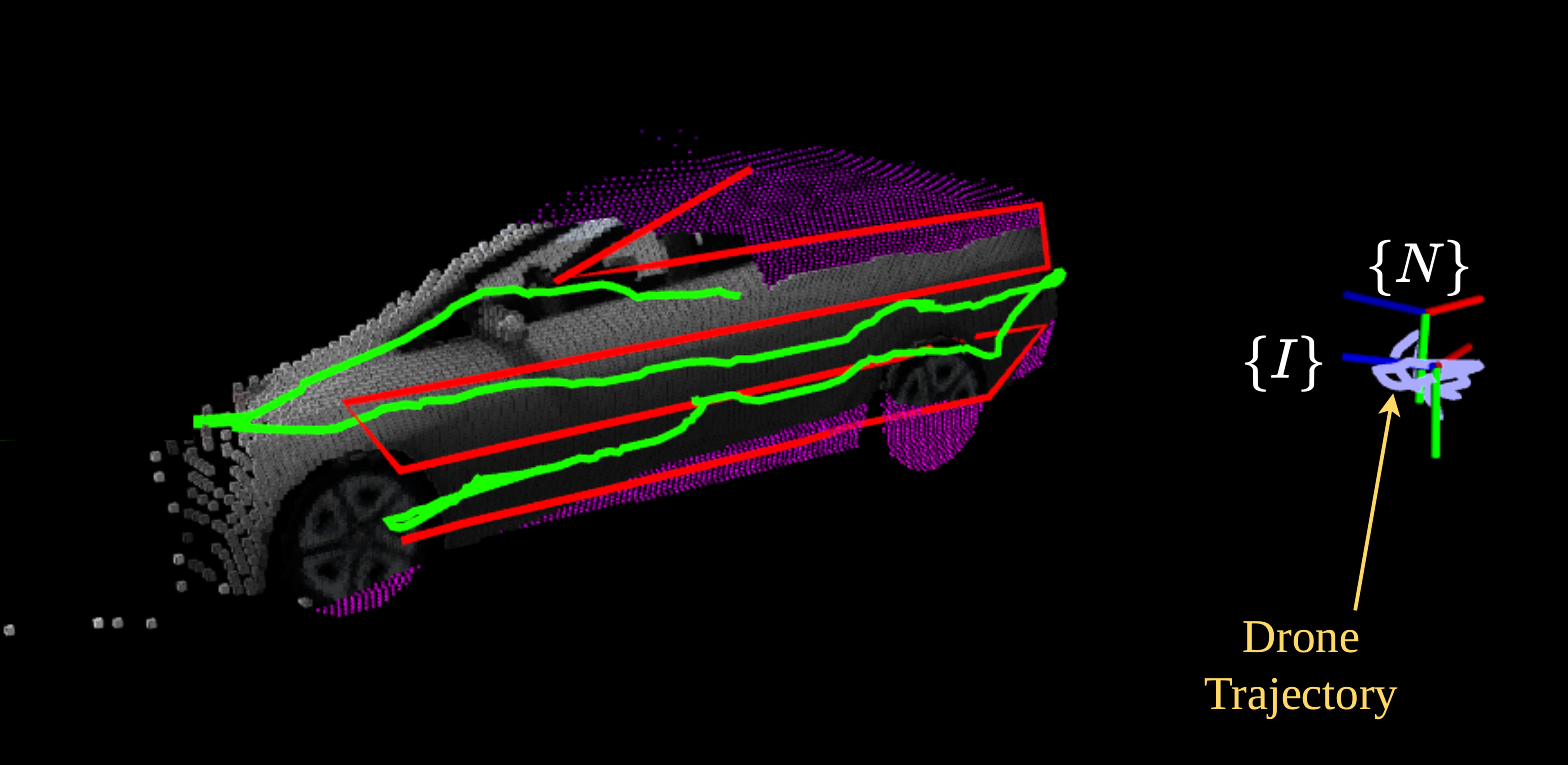}
        \caption{}
        \label{fig:position_err_ff}
    \end{subfigure}
    \hfill
    \begin{subfigure}[b]{0.19\textwidth}
        \centering
        \includegraphics[width=\linewidth,height=2.5cm,trim=6 10 6 10,clip]{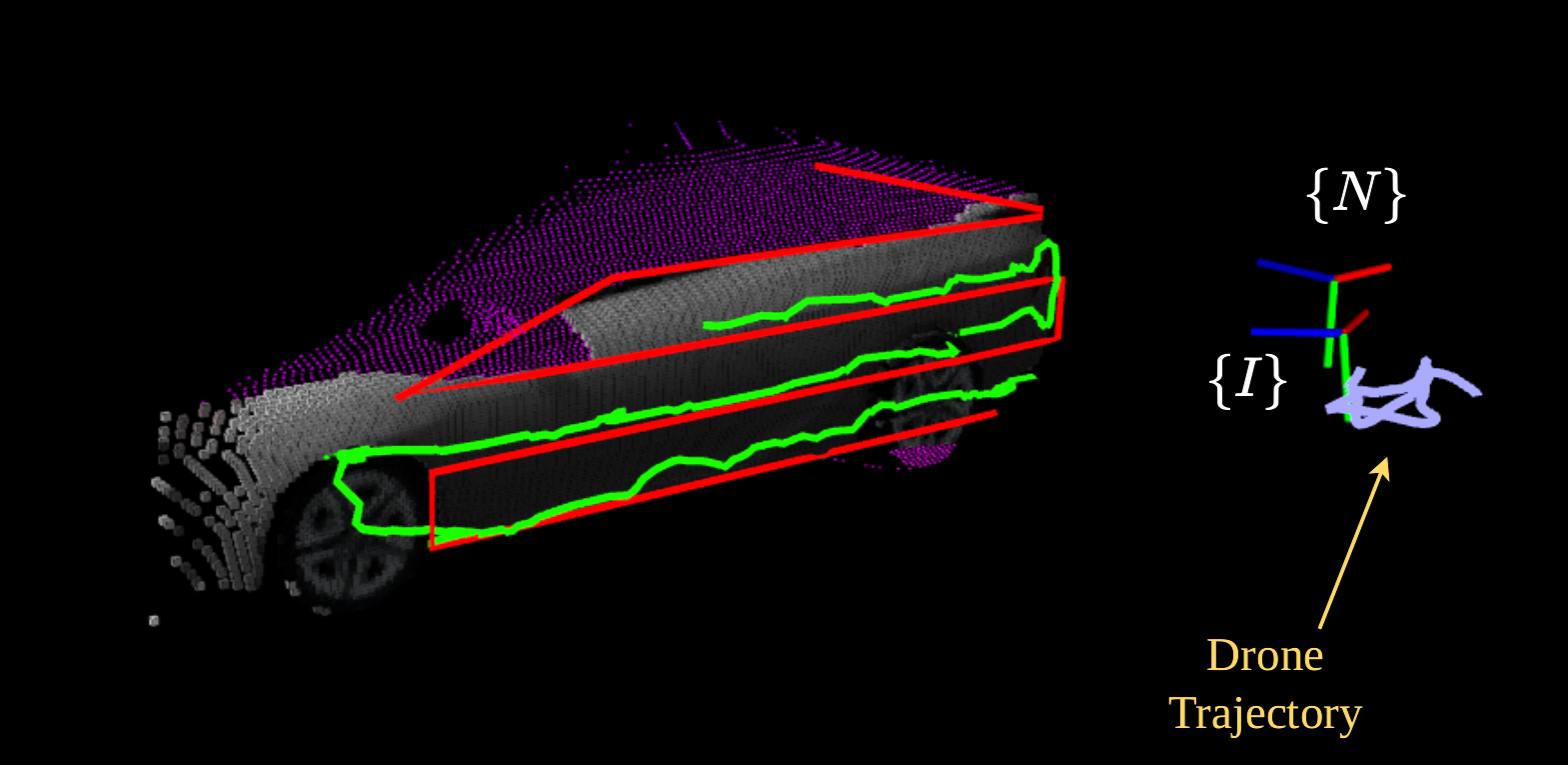}
        \caption{}
        \label{fig:position_err_fb}
    \end{subfigure}
    \caption{Simulation evaluation for car-body surface inspection. The green path is traced by the gimballed camera pointing ray (the $\hat{z}$ axis of $\{I\}$) on the target surface, while the red path is the desired setpoint trajectory. (a) Half-completed sweep. (b) Effect of disturbance with surface-tracking feedback disabled. (c) Effect of disturbance with surface-tracking feedback enabled. (d) GPS noise ($\sigma=1.0\,\mathrm{m}$) without feedback. (e) GPS noise ($\sigma=1.0\,\mathrm{m}$) with feedback.}
    \label{fig:coverage}
\end{figure*}

The partitions need to be sequentially inspected by the gimbal as depicted in Fig.~\ref{fig:pipeline} through the \textit{Gimbal Trajectory Execution} block.
At fixed intervals (e.g., every 2 seconds), a new partition is selected and passed to the \textit{pointing} process.
The gimbal setpoint is computed by mapping the partition center ${}^{N}\mathbf{u}$ with its average depth $d$ and the extrinsic calibration parameters (${}^{G}\mathbf{R}_{N}, {}^{G}\mathbf{t}_{N}$) through Eqs.~\ref{eqn:eqn1} and \ref{eqn:eqn2}, yielding the target position ${}^{G}\mathbf{c}$ in the gimbal base frame.
Then, the desired pan and tilt angles are
\begin{equation}
    \alpha_{\mathrm{pan}} = \arctan\left(\frac{{}^{G}\mathbf{c}_\mathrm{y}}{{}^{G}\mathbf{c}_\mathrm{x}}\right), \qquad
    \alpha_{\mathrm{tilt}} = \arctan\left(\frac{-{}^{G}\mathbf{c}_\mathrm{z}}{\sqrt{{}^{G}\mathbf{c}_\mathrm{x}^2 + {}^{G}\mathbf{c}_\mathrm{y}^2}}\right).
\end{equation}
Finally, the calculated pan and tilt, through the \textit{Gimbal Interface} and PX4, are applied to the gimbal.
The gimbal-mounted camera image is then processed by the \textit{Micro Detector} to detect the target artifact.
\rev{This block is also implemented as a wrapper around YOLO-based detection models, and the checkpoint is provided as a runtime parameter.}

\section{\rev{Implementation \& Results}}
\label{sec:implementation}
The aerial inspection pipeline is implemented in ROS 2 for accessibility and was validated in simulation and real-world experiments which will be discussed in this section.

\subsection{Use Cases}
The pipeline was initially developed for two use cases.
The first addresses widespread oak processionary caterpillars in the Netherlands, which pose health risks to humans and animals\rev{~\cite{jans2011eikenprocessierups}}.
Current treatment relies on mass pesticide spraying, which is costly and environmentally harmful.
In collaboration with the municipality of Enschede and the province of Overijssel (funded by Regieorgaan SIA), we investigated autonomous aerial detection of caterpillars and eggs to support targeted treatment.
\rev{Individual caterpillars are typically less than 40\,mm in size, and their eggs are even smaller at around 1--2\,mm, necessitating high-detail imaging.
This use case fits our definition of aerial micro-inspection: fine-detail images must be captured and analyzed while flying close to the tree poses collision risks with branches and leaves, and the drone is subject to outdoor disturbances and localization inaccuracies, preventing steady focusing on tree trunks from meters away.}
Figures \ref{fig:impl-a} and \ref{fig:impl-b} show infected trees, while Figs.~\ref{fig:impl-e} and \ref{fig:impl-f} show controlled mockup-tree experiments, related to this use case.

The second use case targets monitoring in Dutch flower greenhouses.
Whiteflies can severely damage crops, are very small (1--2 mm), and are difficult to monitor manually at greenhouse scale.
\rev{Autonomous aerial inspection for these whiteflies is also relevant to aerial micro-inspection: the drone must avoid damaging plants due to downwash, prevent collisions with greenhouse structures, minimize flight path length for efficiency, and cope with localization inaccuracies in GPS-denied environments.}
As a step toward automated monitoring and treatment, we evaluated the same aerial inspection workflow in this setting.
Figures \ref{fig:impl-c} and \ref{fig:impl-d} show our experiments for testing our pipeline in a flower greenhouse with sticky traps that catch the flies.

\rev{Another use-case is industrial quality control such as visual inspection of vehicle body surfaces for production defects~\cite{chang2019mobile}.
Vehicle inspection is investigated in simulation here, mainly to demonstrate the transferability of the proposed pipeline to different applications.}

\subsection{Simulation}

We used Gazebo as the simulation environment.
To match our use cases, we created custom worlds including oak-tree and car inspection scenes.
Using PX4 Software-In-The-Loop (SITL), we extended the \texttt{X500} \rev{quadrotor} model with a gimbal-mounted inspection camera ($12^{\circ}$ horizontal FOV (HFoV)) and an RGB-D navigation camera ($120^{\circ}$ HFoV).
\rev{\textit{Micro XRCE-DDS} has been used to bridge ROS~2 and PX4 communication.}
Because required gimbal control/status messages were unavailable over Micro XRCE-DDS in our setup, we added a dedicated \textit{Gimbal Interface} node using MAVLink over UDP (simulation) or serial (real setup).
The simulated scenario involves the drone taking off and moving to a waypoint in the vicinity of the target such that its heading is broadly directed towards the target. 
Our pipeline then acquires the target surface online and performs inspection.
Figures \ref{fig:impl-g} and \ref{fig:impl-h} show the simulated environments and the aerial robot for the tree trunk and car inspection use cases, respectively.

Coverage is a common inspection-quality metric~\cite{chen2023flight}.
In simulation, we reconstruct the target 3D surface (purple point cloud in Fig.~\ref{fig:coveage_vis}) from ground-truth depth and segmentation in the navigation-camera view.
Using the inspection camera's ground-truth pose, we project its FOV and center onto that surface.
In Fig.~\ref{fig:coveage_vis}, the center trajectory of the gimbal FOV is shown in green, and visited surface regions are colored with RGB values.
Coverage is defined as the ratio between visited surface and total target surface.
We configure our pipeline to create partitions with $0\%$ overlap such that a successful inspection should yield a coverage of $100\%$. 

\begin{figure}[!t]
    \centering
    \begin{subfigure}[b]{0.35\textwidth}
        \centering
        \includegraphics[width=\linewidth,height=4.6cm]{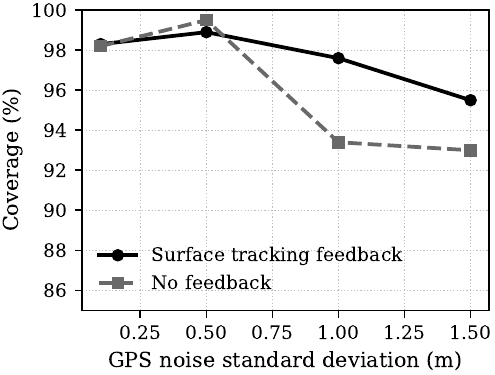}
    \end{subfigure}
    \caption{Effect of GPS-noise-induced drone wobble on target-surface coverage.}
    \label{fig:coverage_vs_gps_noise}
\end{figure}

Flight disturbances directly affect gimbal trajectory execution.
Figure~\ref{fig:disturbance_ff} shows the effect of a disturbance equal to ${}^{N}\mathbf{F}_\mathrm{dist}=[-50, -100, 0]$ Newtons applied to the drone at a specified moment. 
Without feedback, the gimbal trajectory deviates from the setpoint path (defined by partition centers).
With the proposed surface-tracking feedback, it recovers quickly and follows the setpoint path (Fig.~\ref{fig:disturbance_fb}).

Imperfect position holding during the flight is another challenge for micro-inspection.
This behaviour often stems from low-precision localization or control issues.
\rev{We simulated this effect by adding gaussian noise to the drone's simulated GPS sensor and repeating inspection experiment with different standard deviations of noise.}
Figure~\ref{fig:coverage_vs_gps_noise} shows the resulting coverage for each noise level.
Figures~\ref{fig:position_err_ff} and \ref{fig:position_err_fb} compare gimbal trajectory tracking without and with feedback, respectively, under GPS noise with a standard deviation of $1.0\,\mathrm{m}$.

\subsection{Real Setup}
\begin{figure}[!t]
    \centering
    \includegraphics[width=\linewidth,height=5.6cm]{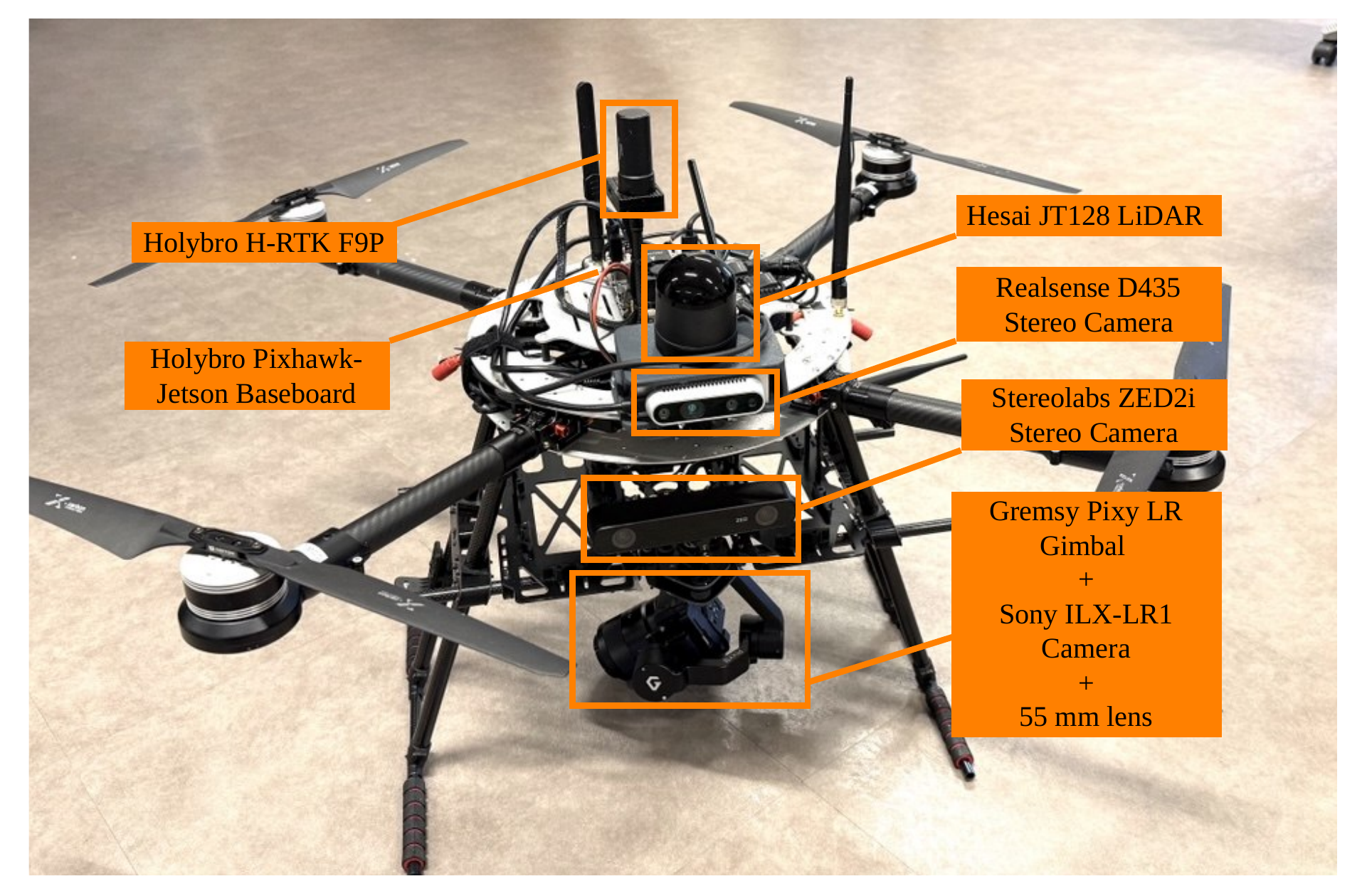}
    \caption{The drone platform used for real-world experiments.}
    \label{fig:sarax2}
\end{figure}

For our experiments, we used a custom quadrotor, based on our open-source aerial robot design, namely SARAX~\cite{alharbat2024sarax} \rev{as displayed in Fig.~\ref{fig:sarax2}.}
We used a \rev{\textit{Holybro Pixhawk Jetson Baseboard}} containing a \textit{Pixhawk} flight controller and a \textit{Jetson Orin Nano} as the companion computer.
\rev{To support stable flight in position hold mode, a \emph{Holybro H-RTK F9P} module was added to the flight stack.}
As the navigation camera, we used a \textit{StereoLabs ZED 2} RGBD camera featuring a 110° HFoV and as the inspection camera we deployed a \textit{Gremsy Pixy LR} gimbal with a \textit{Sony ILX-LR1} camera mounted on it.
The inspection camera was equipped with a 55 mm lens and a 4 times digital zoom, resulting in an effective 9.2° HFoV, yielding an HD resolution image.
\rev{Note that not all the components in the displayed setup were used for this work.}

\begin{figure}[!t]
    \centering
    \begin{subfigure}[b]{0.90\columnwidth}
        \centering
        \includegraphics[width=\linewidth,height=4.0cm]{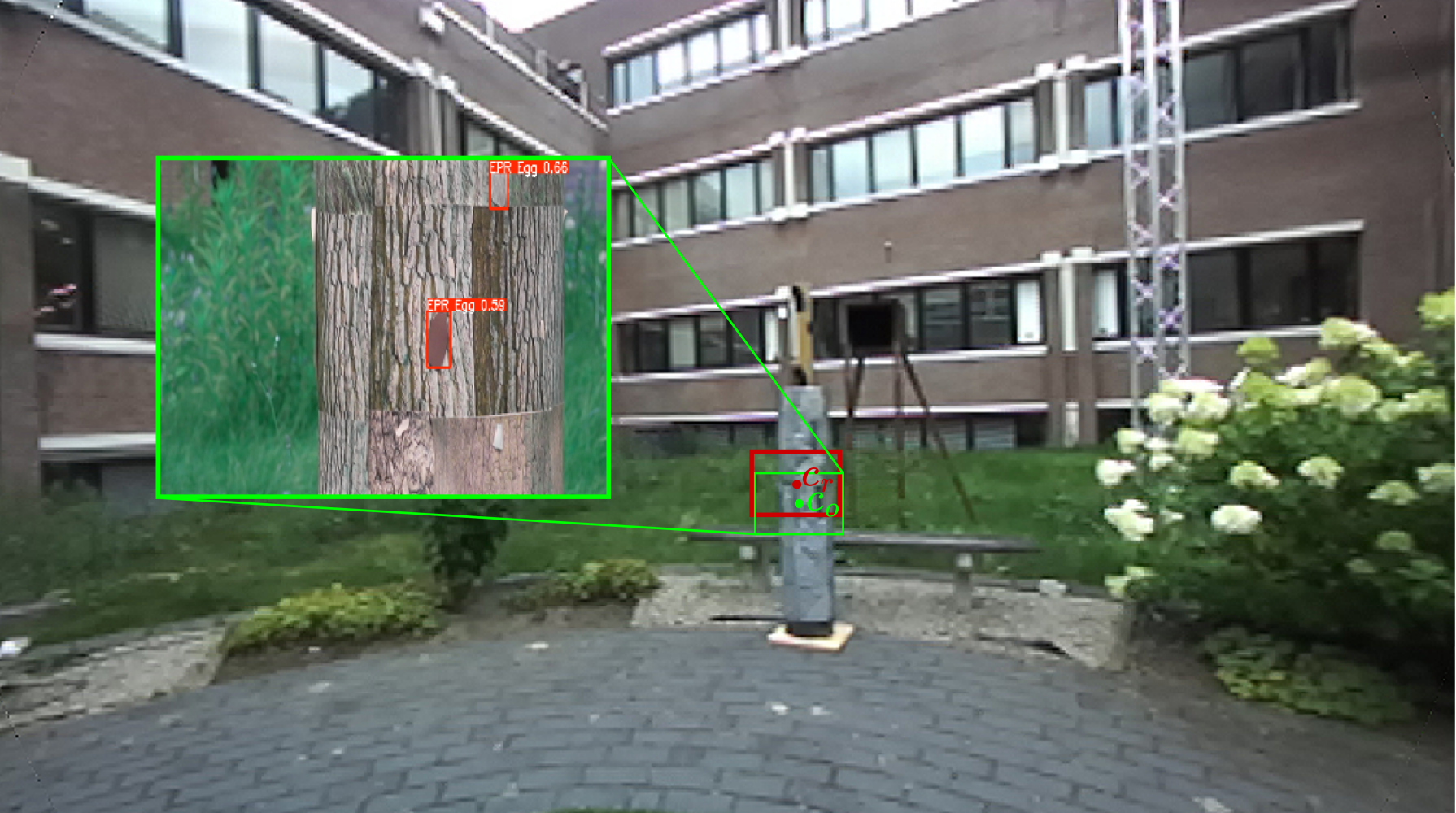}
        \caption{}
        \label{fig:epr_real_err}
    \end{subfigure}
    \hfill
    \begin{subfigure}[b]{0.90\columnwidth}
        \centering
        \includegraphics[width=\linewidth,height=4.0cm]{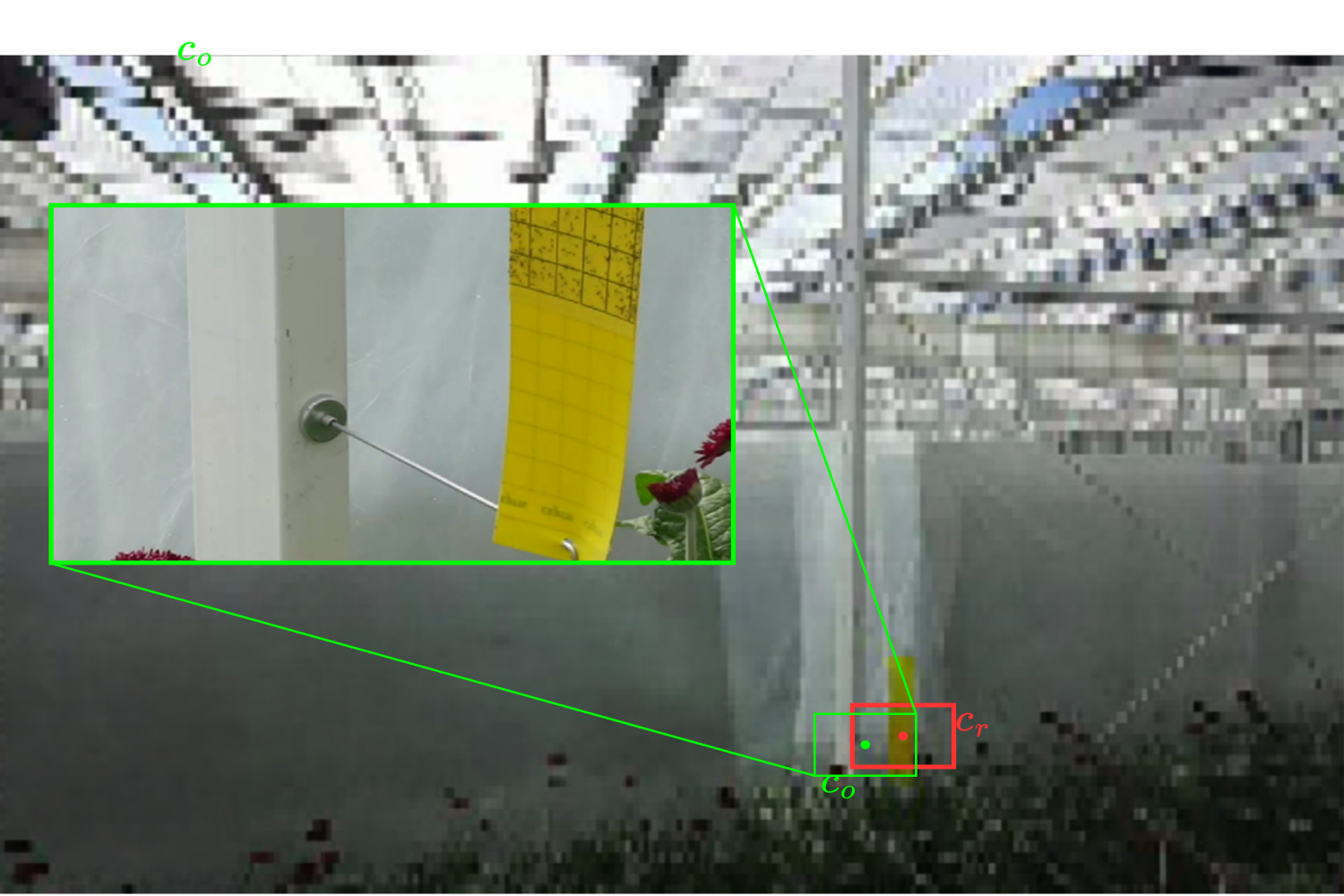}
        \caption{}
        \label{fig:greenhouse_real_err}
    \end{subfigure}
    \caption{Gimbal pointing accuracy assessment in real-world experiments.}
    \label{fig:real_err}
\end{figure}

To investigate the pipeline in real-world conditions, coverage evaluation is less direct because ground-truth surface geometry and gimbal pose are unavailable.
As shown in Fig.~\ref{fig:real_err}, we compared the inspection-camera FOV projected in the navigation image (green, manually annotated) against the intended partition (red) for each partition in a full sweep.
Let the inspection FOV center be $\mathbf{c}_\mathrm{o}=(u_\mathrm{o},v_\mathrm{o})$ and the setpoint partition center be $\mathbf{c}_\mathrm{r}=(u_\mathrm{r},v_\mathrm{r})$. The angular error is

\[
\theta =
\cos^{-1}
\left(
\frac{\mathbf{d}_\mathrm{o}^\top \mathbf{d}_\mathrm{r}}
{\|\mathbf{d}_\mathrm{o}\|\,\|\mathbf{d}_\mathrm{r}\|}
\right),
\]
where
\[
\mathbf{d}_\mathrm{o} =
\begin{bmatrix}
(\bar{u}_\mathrm{o} - c_\mathrm{x})/f_\mathrm{x}\\
(\bar{v}_\mathrm{o} - c_\mathrm{y})/f_\mathrm{y}\\
1
\end{bmatrix},
\qquad
\mathbf{d}_\mathrm{r} =
\begin{bmatrix}
(\bar{u}_\mathrm{r} - c_\mathrm{x})/f_\mathrm{x}\\
(\bar{v}_\mathrm{r} - c_\mathrm{y})/f_\mathrm{y}\\
1
\end{bmatrix}.
\]
where $f_\mathrm{x}, f_\mathrm{y}, c_\mathrm{x},$ and $c_\mathrm{y}$ are navigation-camera calibration parameters.
For the inspection of the mockup tree trunk, the average angular error was $2.17^\circ$.
In the greenhouse sticky-trap experiment, the average error was $3.34^\circ$.

\section{Discussion}
\label{sec:discussion}
The simulation results confirm that the surface-tracking feedback loop substantially improves robustness under flight disturbances and localization noise.
Coverage degrades only mildly with our pipeline despite increased platform wobble, primarily because the feedback mechanism compensates for disturbances in real time.
By contrast, open-loop execution of desired gimbal orientations (as in several related works~\cite{haneda2025viewpoint,wu2025ptz}) shows a stronger coverage drop as noise increases.
The position error comparison in Figs.~\ref{fig:position_err_ff} and \ref{fig:position_err_fb} further illustrates that GPS-induced drift degrades gimbal tracking in open loop, while feedback largely mitigates this effect.

In real-world experiments, the $2.17^\circ$ average angular error achieved in the mockup tree inspection was sufficient for reliable caterpillar detection.
The higher $3.34^\circ$ error in the greenhouse sticky-trap case, combined with the current zoom level, was enough to have high quality pictures of insects on the trap but was insufficient to distinguish whiteflies in comparison to other insects.
Spatially varying errors likely stem from modeling and calibration simplifications; in particular, the assumption of perfect alignment between the inspection camera and gimbal end-effector is only approximate.

\rev{We also observed that the gimbal's internal zero pan/tilt reference can shift between power cycles, introducing a bias that changes at every start-up unless recalibrated.
This effect likely originates from IMU-encoder fusion drift or steady-state error in the gimbal controller.
Repeated trials with a stationary drone showed pointing errors of up to $1.3^\circ$ between different power cycles.
The resulting pixel displacement depends on the zoom level and can be approximated as $\Delta p \approx f \times \tan(\Delta \theta)$.
At the base 55\,mm lens ($f \approx 1972$ px), this yields $\Delta p \approx 45$ px (3.5\% of inspection camera's image width), which is negligible.
However, with $10\times$ zoom, the error grows to $\Delta p \approx 448$ px (${\approx}35\%$ of inspection camera's image width), which can cause a substantial portion of the intended inspection partition to be missed.
A practical short-term mitigation is to increase partition overlap. As a more robust solution, we are considering online cross-camera feature matching for real-time extrinsic refinement.}

\section{Conclusion}
\label{sec:conclusion}
This paper presented \texttt{aerial\_micro\_inspection}, a
modular and open-source dual-camera inspection pipeline for detecting \rev{small} objects.
The proposed framework combines a wide-FOV navigation camera for on-the-spot surface acquisition and partitioning with a narrow-FOV gimballed inspection camera for high-detail visual sensing. 
In simulation, the results showed that integrating surface-tracking feedback into gimbal control improves robustness against platform disturbances and localization noise, leading to more reliable target coverage than open-loop pointing. 
Real-world experiments on tree and greenhouse scenarios further demonstrated that the pipeline can achieve practical pointing accuracy for fine inspection tasks while highlighting key hardware and modeling limitations that affect performance at higher zoom levels.

Future work will focus on both enhancing the model accuracy and operational scalability. 
First, the current kinematic model will be extended to capture additional gimbal-camera effects, such as non-ideal rotation centers, small mechanical offsets, and boot-dependent orientation biases, to reduce pointing errors. 
Second, calibration will be upgraded from an offline target-based procedure toward online, targetless methods that can re-estimate extrinsics at every boot from
cross-camera feature correspondences. 
Third, we will expand the planning layer from local partition sweeping to hierarchical viewpoint generation for larger structures, where multiple wide-view acquisitions must be stitched and scheduled to inspect surfaces that do not fit in a single peripheral view.
Finally, the same modular workflow will be validated in broader
domains, including infrastructure assets, public transport and
vehicle body inspection, and power-line corridor monitoring.



\appendix
\subsection{Kinematics Between Navigation and Inspection Cameras}
\label{app:kinematics}

Let ${}^{N}\mathbf{u}=[u,\,v]^T$ denote the pixel location of the target detected in the navigation image. 
Using the pinhole model \rev{and assuming undistorted images}, the corresponding 3D point in $\{N\}$ is
\begin{equation}
    {}^{N}\mathbf{c} = d \cdot \mathbf{K}_\mathrm{nav}^{-1}{}^{N}\tilde{\mathbf{u}}, \qquad
    {}^{N}\tilde{\mathbf{u}} = [u,\,v,\,1]^T,
    \label{eqn:eqn1}
\end{equation}
where $\mathbf{K}_\mathrm{nav}$ is the intrinsic calibration matrix of the navigation camera and $d$ is the distance of the target from the navigation camera along optical axis.

We model the rigid transform between $\{N\}$ and $\{G\}$ as a constant extrinsic:
\begin{equation}
{}^{G}\mathbf{T}_{N} =
\begin{bmatrix}
{}^{G}\mathbf{R}_{N} & {}^{G}\mathbf{t}_{N}\\
\mathbf{0} & 1
\end{bmatrix},
\end{equation}
\rev{transforming the same point $\mathbf{c}$, but in the gimbal base frame $\{G\}$:}
\begin{equation}
    {}^{G}\mathbf{c} = {}^{G}\mathbf{R}_{N}{}^{N}\mathbf{c} + {}^{G}\mathbf{t}_{N}.
    \label{eqn:eqn2}
\end{equation}

We can obtain the target point in the gimbal's instantaneous frame $\{G_t\}$ by applying the gimbal's current orientation ${}^{G_t}\mathbf{R}_{G}$ to ${}^{G}\mathbf{c}$. \rev{Since the gimbals might not perfectly rotate around their center of rotation, an induced translation of ${}^{G_t}\mathbf{t}_{G}$ is assumed:}
\begin{equation}
    {}^{G_t}\mathbf{c} = {}^{G_t}\mathbf{R}_{G}\,{}^{G}\mathbf{c} + {}^{G_t}\mathbf{t}_{G}.
\end{equation}

\rev{In reality, the gimbal's instantaneous frame $\{G_t\}$ might slightly differ from the camera's frame $\{C\}$, due to mechanical imperfections with an angle of $\delta\boldsymbol{\alpha}$ around an unknown axis $\mathbf{n}$ and some translation $\delta\mathbf{t}$.} 
Thus, the transformation from $\{G_t\}$ to the inspection camera's \rev{optical} frame $\{I\}$ can be broken down to:
\begin{equation}
{}^{I}\mathbf{T}_{G_t}
=
{}^{I}\mathbf{T}_{C}\;
{}^{C}\mathbf{T}_{G_t},
\qquad
{}^{C}\mathbf{T}_{G_t} \approx
\begin{bmatrix}
\mathbf{R}(\delta\boldsymbol{\alpha}, \mathbf{n}) & \delta\mathbf{t}\\
\mathbf{0} & 1
\end{bmatrix},
\end{equation}
where the ${}^{I}\mathbf{T}_{C}$ is a re-axis rotation:

\begin{equation}
{}^{I}\mathbf{T}_{C} =
\begin{bmatrix}
\mathbf{R}_{C} & \mathbf{0}\\
\mathbf{0} & 1
\end{bmatrix},
\qquad
\mathbf{R}_{C} =
\begin{bmatrix}
0 & 1 & 0\\
0 & 0 & 1\\
1 & 0 & 0
\end{bmatrix}.
\end{equation}
The target point in the inspection camera frame will be
\begin{equation}
    {}^{I}\mathbf{c} = {}^{I}\mathbf{T}_{G_t}{}^{G_t}\mathbf{c}.
\end{equation}    

Finally, the pixel location of the target in the inspection image is
\begin{equation}
    {}^{I}\mathbf{u} = \mathbf{K}_\mathrm{ins}\frac{{}^{I}\mathbf{c}}{{}^{I}\mathbf{c}_\mathrm{z}},
\end{equation}
where $\mathbf{K}_\mathrm{ins}$ is the intrinsic calibration matrix of the inspection camera and \rev{${}^{I}\mathbf{c}_\mathrm{z}$ is the third component of ${}^{I}\mathbf{c}$ and serves as the depth value, as visualized in Fig.~\ref{fig:kinematic_model}.}

\bibliographystyle{IEEEtran} 
\bibliography{refs}

\vspace{12pt}

\end{document}